\documentclass[11pt]{article}

\usepackage[final]{acl}
\usepackage{times}
\usepackage{latexsym}

\usepackage[T1]{fontenc}
\usepackage[utf8]{inputenc}

\usepackage{microtype}

\usepackage{inconsolata}

\usepackage{graphicx}

\usepackage{paralist}
\usepackage{amsmath}
\usepackage{amssymb}
\usepackage{tcolorbox}
\tcbuselibrary{skins, listings, breakable}
\usepackage{xcolor}
\usepackage{algorithm}
\usepackage{algpseudocode}
\usepackage{adjustbox}
\usepackage{booktabs}
\usepackage{multirow}
\usepackage{xcolor,colortbl}

\newtcolorbox{findingbox}[1]{colback=gray!6,colframe=gray!60,boxrule=0.5pt,arc=2pt,left=4pt,right=4pt,top=2pt,bottom=2pt,toptitle=1pt,bottomtitle=1pt,fontupper=\small,title={#1},fonttitle=\small\bfseries,coltitle=black,colbacktitle=gray!18}
\usepackage{colortbl}
\usepackage{graphicx}
\usepackage{subcaption}
\usepackage{tabularx}
\usepackage{caption}
\usepackage{array}

\usepackage{listings}
\usepackage{xcolor}
\usepackage{placeins}   % \FloatBarrier before each appendix section keeps its tables and figures with its text

\lstdefinelanguage{json}{
  basicstyle=\ttfamily\small,
  numbers=left,
  numberstyle=\tiny,
  stepnumber=1,
  numbersep=6pt,
  showstringspaces=false,
  breaklines=true,
  frame=single,
  backgroundcolor=\color{gray!5},
  tabsize=2
}

\title{From Single-Actor Environments to Multi-Actor Event Worlds:\\ Can LLMs Be Event World Models?}

\author{
Juntong Ni, Zewen Liu, Bohan Wang, Lige Zhang, Wei Jin
\\
Emory University \\
\texttt{\{firstname, lastname\}@emory.edu}
}

\begin{document}
\maketitle
\begin{abstract}
World models help agents assess the consequences of their actions before acting. Many real systems, however, involve multiple actors who make their own decisions, can evolve even when the agent takes no action, and are only partially observed through event logs. We formalize these systems as \emph{event worlds} and ask whether large language models can predict future observations from the log and the agent's proposed actions. We build complementary benchmarks from human Diplomacy games and real Polymarket trades. Our hierarchical framework compares prediction with and without other actors' actions, then tests multi-step rollouts, predictions for actions never taken, and usefulness to agents. Fine-tuning and more data improve prediction when all actors' recorded actions are given. With only the agent's action, models struggle when other actors strongly affect the outcome, and larger models and longer histories help little. This gap also limits multi-step prediction and agent decisions, with models overestimating plan success. Yet imagined experience and initialization from world models can still improve policy learning. Anticipating other actors' actions remains a central challenge for event world modeling.
\footnote{Code: \href{https://github.com/LingFengGold/EventWorld}{github.com/LingFengGold/EventWorld}}
% \footnote{Code: \href{https://anonymous.4open.science/r/EventWorld-F350/}{anonymous.4open.science/r/EventWorld-F350}}
% \footnote{\url{https://anonymous.4open.science/r/EventWorld-F350/}}.
% \lige{I feel the abstract moves too quickly into defining event worlds without showing the motivation. It might be more accessible to audience like: ``World models enable agents to anticipate action outcomes and plan without interacting with the real environment. Existing LLM-based world models often assume that state transitions depend only on the agent's action. In real-world scenarios, however, the state (evolves naturally?) and also depends on other actors whose future actions are hidden from the decision-making agent.''}
\end{abstract}

\section{Introduction}
\label{sec:intro}

\begin{figure}[t]
\centering
\includegraphics[width=\columnwidth]{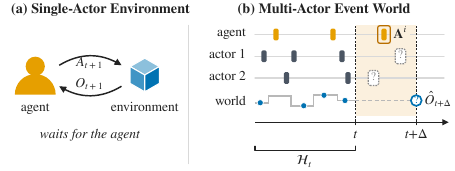}
\caption{From single-actor environments to multi-actor event worlds. $\mathcal{H}_t$, $\textbf{A}$, and $O$ denote the event log, actions, and observations.
% An event world (right) runs on its own clock, is driven by more than one actor, and is known only through the events it emits, its 
% history/log $\mathcal{H}_{t}$. 
% An event world model predicts the observation at $t+\Delta$ from the log and the agent's proposed action, which is outlined, while the other actors' actions after $t$ are unknown.
}
\label{fig:world-model-comparison}
% \vskip -1em
\end{figure}

World models simulate how an environment changes in response to an agent's actions~\citep{hafner2025mastering,hafner2025training}. Such models enable agents to anticipate the consequences of their actions before acting and learn from imagined experience~\citep{ha2018world,gu2024your,li2026word}. Recently, large language models (LLMs) have been used as world models in text, web, and tool environments~\citep{wang2024can,gu2024your,ruan2024identifying,li2026word,zuo2026qwen,li2026bridging}. These studies mainly use environments such as ALFWorld~\citep{shridhar2020alfworld}, WebShop~\citep{yao2022webshop}, and StableToolBench~\citep{guo2024stabletoolbench}.
% In these benchmarks, the environment advances through the agent's actions and returns an observation after each step, so the agent's choices can be tested by running the environment. 
% These environments advance only through the agent's actions and return an observation after every step (Figure~\ref{fig:world-model-comparison}a).
% In these \emph{single-actor environments}, the agent is the only decision-making participant, or \emph{actor}. 
In these environments, the agent is the only decision maker, or \emph{actor}. We thus refer to them as \emph{single-actor environments}.
They advance only through the agent's actions and return an observation after each step (Figure~\ref{fig:world-model-comparison}a).

% \bohan{Considering other participants’ actions alone may not clearly distinguish this work from existing world models. I suggest emphasizing the formulation of event world modeling and the systematic evaluation of LLMs in this setting, while clarifying how these contributions differ from prior multi-agent world model studies.}

% \wei{However, those worlds are limited....(to motivate our event world; make it clear and explicit)}
% However, many real systems continue to evolve while the agent waits, and what happens next also depends on decisions made by other participants. 
However, many real systems evolve through the decisions of multiple actors and are only partially observed through event logs.
% In healthcare, a doctor chooses a treatment, but its outcome also depends on whether the patient follows it. In a market, a trader decides to buy, but what happens to the price also depends on whether other traders buy or sell. In an online auction, a bidder chooses how much to offer, but whether they win depends on what other bidders do. Clinical records, trade logs, and bid histories capture participants' observed actions and changes in the system over time.
% In healthcare, clinical records capture a doctor's prescription and the patient's condition at each visit, while the illness may progress and the patient may miss doses between visits. In a market, trade logs record executed orders and prices but do not reveal traders' intentions. In an online auction, bid histories show a bidder's offer and whether it remains in the lead as others submit competing bids.
In healthcare, a doctor prescribes medication, but the illness continues to evolve and the patient may miss doses between visits. 
In a market, a trader places an order, but prices also change as other traders buy and sell, while their beliefs may change as news comes.
In an online auction, a bidder makes an offer, but can lose the lead when someone else bids. 
% In real systems, rules may determine the consequences of given actions while leaving the participants' choices unknown. In healthcare, a doctor prescribes medication, a patient may take it regularly or miss doses, and the illness continues to evolve. In a market, traders place orders while others revise their quotes. In a multiplayer game, players make competing choices without knowing one another's intentions. Many people act in these systems at the same time, no one can observe all their choices, and no fixed rule determines what each will do next.
% In each case, multiple participants make their own decisions, and the consequences of one participant's actions depend on what the others do. 
In each system, we focus on one actor, the \emph{agent}, whose decisions we aim to support.

We call such systems \emph{event worlds} (Figure~\ref{fig:world-model-comparison}b). They are driven by more than one actor, can evolve even when the agent takes no action, and are known only through event logs that reveal part of the underlying state. 
An event log is a sequence of time-stamped records of actions and observations.
An \emph{event world model} predicts the observation at a chosen time horizon, given the event log and the agent's proposed actions. The other actors' actions during this interval are unknown to the agent.

An event world model can help an agent assess the consequences of its actions before acting, especially when testing alternatives in the real system is costly or restricted. Event logs record such systems at scale. However, prediction from event logs remains difficult, because they reveal only part of the state and leave the other actors' future actions unknown. Event logs also contain heterogeneous records of actions and observations. 
LLMs have the potential to understand these records. Knowledge from pretraining may also help predict how these systems evolve and how the actors behave. We thus ask: \textbf{Can LLMs be event world models?}

To study this question, we construct two complementary event world benchmarks from human game records of \emph{Diplomacy}~\citep{paquette2019no} and trade logs of \emph{Polymarket}~\citep{wang2026polymarket}. We turn these multi-actor event logs into prediction tasks from one agent's perspective, separating its actions from those of the other actors. Diplomacy's rules engine additionally allows us to check the outcomes of actions never taken and evaluate agents beyond the recorded games. Polymarket tests prediction in a real market whose order book is unobserved and for which no engine is available to check actions never taken.

To evaluate comprehensively whether LLMs can serve as event world models, we propose a hierarchical evaluation framework (Figure~\ref{fig:framework}). Each level asks more of LLMs than the level below it. 
\textbf{First}, we evaluate how well LLMs predict future observations given the event log and every actor's recorded action (Section~\ref{sec:q1}). 
\textbf{Second}, we provide only the agent's action alongside the event log, so the models also have to anticipate what the other actors will do (Section~\ref{sec:q2}).
\textbf{Third}, we test situations not in the event log, such as predicting several steps ahead using the models' own predictions and assessing the consequences of actions never taken (Section~\ref{sec:q3}). \textbf{Finally}, we ask whether LLMs help an agent act by planning, by choosing among candidate actions, or by imagining games for the agent to learn from (Section~\ref{sec:q4}).

% To test the last two levels, we need to check the outcomes of actions beyond those in the log. Logs record events that occurred, so they alone cannot establish what alternative actions would cause or how an agent would perform after making different choices~\citep{oberst2019counterfactual}. We therefore build two complementary benchmarks. \emph{Diplomacy} provides human game records and a rules engine for checking alternative orders and evaluating agents. The engine resolves orders but does not determine human choices. \emph{Polymarket} provides real trade logs without the order book, testing prediction under partial observability. It has no engine for verifying alternative orders.

Our experiments show that 
\textbf{(1)} fine-tuning and more data improve prediction when every actor's actions are given. 
\textbf{(2)} However, given only the agent's own action, predictions remain less accurate, especially when other actors' actions have a large effect on the outcome, and this gap remains even with larger LLMs. 
\textbf{(3)} Rollouts accumulate errors, and fine-tuned LLMs predict the effects of actions never taken far less accurately without the other actors' actions.
\textbf{(4)} The same gap affects agent decisions, with LLMs overestimating plan success and providing less help in choosing actions when only the agent's action is known. 
\textbf{(5)} Nevertheless, LLMs as event world models support policy learning through imagined experience and world-model initialization. 
These findings highlight anticipating the other actors' actions as the central open problem of event world modeling.

Our main contributions are as follows.
\begin{compactenum}[(a)]
    \item We define event worlds and event world models, providing a unified formulation for predicting future observations from event logs and an agent's proposed actions in partially observed systems with multiple actors.

    \item We construct two complementary benchmarks and propose a hierarchical evaluation framework covering prediction with and without the other actors' actions, simulation of situations not in the event log, and usefulness to agents.

    \item We systematically evaluate prompted and fine-tuned LLMs, identifying their capabilities and limitations as event world models and providing empirical insights to guide future research.
\end{compactenum}

% Short main-text version; the full discussion with the comparison table is in 2related.tex (Appendix, app:related).
% Citation keys are listed with exact titles at the top of 1introduction.tex and 2related.tex.
\section{Related Work}
\label{sec:related}

% LLMs predict the next state of text, web, and tool environments and help agents plan or train~\citep{wang2024can,gu2024your,li2026word,zuo2026qwen,chu2026agentic}. These models treat the world as waiting for the agent, even when trained on real software~\citep{zuo2026qwen}. Video, game, robot, driving, and traffic models predict frames or scene states from actions~\citep{ha2018world,hafner2025mastering,bruce2024genie,hu2023gaia,montali2023waymo,wu2024smart}. They see the scene densely at a fixed rate, whereas an event world emits sparse events over a partly hidden state. Agent societies and Diplomacy agents model actors inside designed or given worlds~\citep{park2023generative,paquette2019no,meta2022human}, and multi-agent benchmarks control or run every actor in one environment, so they come without records of independent actors to learn from~\citep{guo2024large,zhu2025multiagentbench}. Point processes and LLM forecasters predict what comes next in event logs or how real questions resolve~\citep{mei2017neural,shi2023language,halawi2024approaching,yu2026building}, not what the agent's action changes, and simulators learned from market and clinical logs condition on actions~\citep{coletta2022learning,li2025mars,mu2026ehrworld} but fold the other actors into the world or omit them. Appendix~\ref{app:related} compares these lines of work in full (Table~\ref{tab:related}).

LLM world models predict future observations and support planning and policy learning in text, web, and tool environments~\citep{wang2024can,gu2024your,li2026word,zuo2026qwen,chu2026agentic}. We extend their evaluation from single-actor interaction to multi-actor event logs. 
Visual world models typically predict frames or trajectories sampled at regular intervals~\citep{ha2018world,hafner2025mastering,bruce2024genie,hu2023gaia,montali2023waymo,wu2024smart}, whereas event logs may record actions and observations at irregular times. 
Many multi-agent benchmarks generate interactions using LLM agents in predefined environments~\citep{guo2024large,zhu2025multiagentbench}, while our prediction tasks use human game records and real market logs. 
Event forecasting predicts future events from historical records~\citep{shi2023language,halawi2024approaching,yu2026building}. Our task additionally conditions on the agent's proposed action. 
Market and clinical simulators~\citep{coletta2022learning,li2025mars,mu2026ehrworld} are particularly close, predicting outcomes following trades or treatments. We build on these domain-specific studies with a unified formulation and evaluation, examining how other actors' unknown actions affect prediction and agent usefulness. More details are in Appendix~\ref{app:related}.
\section{Problem Definition}
\label{sec:problem}
\label{sec:ewm-def}

% This section formalizes event world modeling. We define the event world and the agent that acts in the event world, introduce the event world model that serves the agent, and derive two tasks whose comparison separates the event world's response from the other actors' choices.
This section formalizes the event world and the agent that acts in the event world, introduces the event world model that serves the agent, and derives two tasks.

\paragraph{Event World.}
An event world is driven by a set of actors $\mathcal{I}$ with $|\mathcal{I}| \ge 2$, can evolve even when the agent takes no action, and is known only through its \emph{event log}. 
Its state $S_t = (O_t, X_t)$ at time $t$ consists of the latest \emph{observation} $O_t$ of the world and a hidden part $X_t$, such as the actors' intentions, unrecorded actions, or the world's internal conditions.
The state evolves as $S_{t+1} \sim P(\cdot \mid S_t, A_{t+1})$, where $A_{t+1}$ are all actors' recorded actions at step $t+1$, possibly none. 
Actions and observations need not occur at every time step, whereas $X_t$ can evolve at every step.
Thus, even when the agent takes no action, the state can evolve through the other actors' actions or through changes in $X_t$ alone.
The event log $\mathcal{H}_t = (A_{\le t}, O_{\le t})$ records the actions and observations up to $t$.

% \bohan{Should we clarify what "runs on its own clock" means in both synchronous Diplomacy and asynchronous Polymarket?}

\paragraph{Agent in an Event World.}
We take the view of one actor $i$, the \emph{agent}, defined as
\begin{equation}
\begin{aligned}
\mathcal{A}&: \mathcal{H}_{t} \to A^i_{t+1}, \\
\mathcal{H}_{t} &= \big(A^i_{\le t},\, A^{-i}_{\le t},\, O_{\le t}\big),
\end{aligned}
\label{eq:agent}
\end{equation}
% \bohan{is $S_{\le t}$ an observation rather than the full underlying state?}
where $A^i_{\le t}$ and $A^{-i}_{\le t}$ are the agent's own actions and the other actors' recorded actions up to $t$. At time $t$, the agent uses $\mathcal{H}_{t}$ to choose its next action $A^i_{t+1}$, which could be empty, meaning that it waits. The other actors choose theirs at the same time.

\paragraph{Event World Model.}
An agent chooses $A^i_{t+1}$ better if it can foresee what its actions lead to. An \emph{event world model} makes this prediction,
\begin{equation}
\mathcal{W}: \big\{\mathcal{H}_{t}, \mathbf{A}^i;\, \Delta\big\} \to \hat{O}_{t+\Delta},
\label{eq:ewm}
\end{equation}
where $\mathbf{A}^i = A^i_{(t,\,t+\Delta]}$ are the actions the agent proposes in the interval $(t, t+\Delta]$, $\Delta$ is a time horizon chosen by the modeler, and $\hat{O}_{t+\Delta}$ is the predicted $O_{t+\Delta}$.

\paragraph{Two Tasks.}
At each \emph{decision point} $t$, we compare two prediction tasks. 
In \emph{Task~S} (agent's action only), the model receives the event log and the agent's proposed actions, as in Eq.~\eqref{eq:ewm}.
In \emph{Task~J} (joint action), it also receives the other actors' recorded future actions, $\mathbf{A}^{-i} = A^{-i}_{(t,\,t+\Delta]}$. 
Task~J tests prediction given the joint action $\mathbf{A} = (\mathbf{A}^i, \mathbf{A}^{-i})$.
For a fixed $\Delta$, the two conditional distributions satisfy
\begin{multline}
P\big(O_{t+\Delta} \mid \mathcal{H}_{t}, \mathbf{A}^i\big) = \\
\sum_{\mathbf{A}^{-i}} P\big(\mathbf{A}^{-i} \mid \mathcal{H}_{t}, \mathbf{A}^i\big)\, P\big(O_{t+\Delta} \mid \mathcal{H}_{t}, \mathbf{A}\big).
\label{eq:tasks}
\end{multline}
The first factor describes how the other actors may act, and the second describes the world's response to the joint action.
The gap between the tasks shows how much knowing the other actors' actions improves prediction.

\section{Event World Benchmarks}
\label{sec:worlds}

% \juntong{show our effort, make it significant}
To evaluate how well LLMs predict future observations and support agent decisions, we build two complementary benchmarks from multi-actor event logs (Table~\ref{tab:benchmarks}). Both support prediction of recorded outcomes. Diplomacy additionally provides a rules engine for checking actions never taken and evaluating agents. Polymarket has no such engine, but it tests prediction in a real market where many traders act at any moment.
% \lige{Both Diplomacy and Polymarket are still largely driven by actors' actions. It seems that the world evolves mainly in response to these actions, rather than independently of them. However, my understanding of an event world is that the world somehow evolves on its own, but can also be influenced by actors' actions. (I feel the key distinction is between "a world that evolves independently of the focal agent's actions" and "a world that has its own autonomous state dynamics", even in the absence of any actors' actions. My understanding of "runs on its own clock" is closer to the latter.)}

\begin{table}[t]
\centering
\footnotesize
\setlength{\tabcolsep}{3pt}
\begin{tabularx}{\columnwidth}{@{}l>{\raggedright\arraybackslash}X>{\raggedright\arraybackslash}X@{}}
\toprule
& Diplomacy & Polymarket \\
\midrule
World type & board game & market \\
Actors $\mathcal{I}$ & 2 or 7 players & anonymous traders \\
Observation $O$ & the board & trading statistics \\
Hidden part $X$ & players' intentions & order book, beliefs \\
Actions $A$ & move, hold, support & buy or sell \\
Horizon $\Delta$ & one phase & one hour \\
Simulator & rules engine & none \\
\bottomrule
\end{tabularx}
\caption{The two event world benchmarks.}
\label{tab:benchmarks}
% \vskip -1em
\end{table}

\paragraph{Diplomacy.}
Diplomacy is a board game in which seven players move armies and fleets on a map of Europe to capture supply centers. In each round, called a phase, all players write their orders in secret, and the orders are carried out together. Units move and fight in movement phases, beaten units withdraw in retreat phases, and players add or remove units in adjustment phases. 
The players' intentions, the hidden part $X_t$, can evolve on their own. 
Given the current board and all players' orders, the rules determine the next board $O_{t+1}$.

\paragraph{Polymarket.}
Polymarket is a market for betting on future events, where a share's price, between 0 and 1, reads as the chance that the event happens. Traders buy and sell at any moment, and each trade is filled by the waiting offers in the order book, which our trade logs do not show. These offers and the traders' beliefs, the hidden part $X_t$, may change between recorded trades as news arrives. The same trade can therefore move the price by different amounts, so the price stays uncertain even when every trader's orders are known.

\paragraph{Benchmark Construction.}
We turn the raw records into event world benchmarks in two steps. \textbf{First}, we prepare and verify the records. For Diplomacy, we replay the 33,279 games without negotiation from webDiplomacy~\citep{paquette2019no} with the rules engine and keep the 99.7\% that it reproduces exactly, so the engine can be trusted for any orders. For Polymarket, we take the on-chain trade records~\citep{wang2026polymarket}, keep the 4,176 markets traded for at least 30 days with at least 5,000 trades, and merge their trades into 79M orders by trader. 
\textbf{Second}, we turn each event log, produced by all actors together, into samples from one agent's view. At each decision point $t$, we choose the agent, rebuild the history $\mathcal{H}_t$ known at $t$, and split the actions in $(t, t+\Delta]$ into the agent's $\mathbf{A}^i$ and the other actors' $\mathbf{A}^{-i}$. Task~S receives $\mathbf{A}^i$, Task~J also $\mathbf{A}^{-i}$, and both predict the same recorded $O_{t+\Delta}$. The change from $O_{t}$ to $O_{t+\Delta}$ is a \emph{transition}.  In Diplomacy, we select two players at each phase and construct a separate sample from each player's perspective. In Polymarket, decision points are the moments when a trader places a large order, with this trader as the agent, plus random moments without an agent action. More details about benchmark construction are in Appendix~\ref{app:bench}.
This yields 940k transitions from seven-player Diplomacy games and 1.02M one-hour transitions from Polymarket. 
% A transition is high-interaction if its outcome depends on how several actors' actions combine: movement phases in Diplomacy and hours with at least two orders from other traders in Polymarket.
We split the data by game or market event, with statistics and details in Appendix~\ref{app:data}.

% \paragraph{Data Splits.}
% We split the 21,832 seven-player Diplomacy games into training (19,721), validation (1,212), and test (899) sets, and keep the 11,362 two-player games to evaluate robustness to a different number of players. For Polymarket, markets of events that started before October 2025 form the training (2,170 markets) and test (258) sets. We use the 890 markets of events that started in 2026 to evaluate robustness to later markets. Because the closed models are costly, all models are compared, unless stated otherwise, on shared subsets of the test sets, 1,000 transitions in Diplomacy (2,000 Task~S samples) and 1,333 decision points in Polymarket (333 without an agent action). To examine settings where interactions among actors are more likely, we label movement phases in Diplomacy and windows with at least two orders from other traders in Polymarket as \emph{high-interaction} and \emph{low-interaction} otherwise.

\begin{table*}[t]
\centering
\small
\setlength{\tabcolsep}{4pt}
\begin{tabular}{@{}llcccccccc@{}}
\toprule
& & \multicolumn{4}{c}{Diplomacy} & \multicolumn{4}{c}{Polymarket} \\
\cmidrule(lr){3-6}\cmidrule(l){7-10}
Setting & Model & EM & EM$_\mathrm{high}$ & EM$_\mathrm{low}$ & F1 & Acc & Acc$_\mathrm{high}$ & Acc$_\mathrm{low}$ & CRPS$\downarrow$ \\
\midrule
\multicolumn{10}{c}{\textit{Task~J (joint action)}} \\
\midrule
Zero-shot & GPT-6-Luna & 29.1 & 3.4 & 55.2 & 90.8 & 43.0 & 36.8 & 48.0 & 0.42 \\
 & GPT-6-Sol & 40.5 & 10.1 & 71.4 & 96.3 & 54.6 & 48.7 & 59.5 & 0.55 \\
\cmidrule(l){2-10}
SFT & Qwen3-1.7B & 59.7 & 23.6 & 96.4 & 96.9 & \textbf{66.1} & \textbf{62.5} & 69.0 & \textbf{0.29} \\
 & Qwen3-8B & \textbf{65.1} & \textbf{30.8} & \textbf{100.0} & \textbf{97.7} & 65.5 & 60.5 & \textbf{69.5} & \textbf{0.29} \\
\cmidrule(l){2-10}
Baseline & No change & 1.7 & 1.0 & 2.4 & 81.5 & 17.8 & 2.0 & 30.6 & 0.37 \\
 & Orders applied literally & 51.1 & 4.8 & 98.2 & 92.1 & -- & -- & -- & -- \\
 & LightGBM & -- & -- & -- & -- & 68.8 & 67.1 & 70.2 & 0.30 \\
\midrule
\multicolumn{10}{c}{\textit{Task~S (own action only)}} \\
\midrule
Zero-shot & GPT-6-Luna & 6.7 & 1.0 & 12.5 & 82.6 & 46.1 & 40.3 & 50.9 & 0.48 \\
 & GPT-6-Sol & 7.5 & 1.4 & 13.7 & 83.3 & 48.5 & 42.3 & 53.5 & 0.47 \\
\cmidrule(l){2-10}
SFT & Qwen3-1.7B & 20.1 & 1.6 & 38.8 & 84.1 & 59.9 & 55.2 & \textbf{63.7} & \textbf{0.31} \\
 & Qwen3-8B & \textbf{20.8} & \textbf{1.8} & \textbf{40.0} & \textbf{84.3} & \textbf{60.2} & \textbf{55.9} & \textbf{63.7} & 0.32 \\
\cmidrule(l){2-10}
Baseline & No change & 1.7 & 1.0 & 2.4 & 81.5 & 17.8 & 2.0 & 30.6 & 0.37 \\
 & Own orders applied literally & 7.7 & 1.9 & 13.6 & 83.0 & -- & -- & -- & -- \\
 & LightGBM & -- & -- & -- & -- & 61.4 & 58.2 & 63.9 & 0.31 \\
\bottomrule
\end{tabular}
\caption{Single-step prediction (\%, CRPS in cents) on decision points shared by all models, with every actor's action given (Task~J, Q1) or only the agent's (Task~S, Q2). EM$_\mathrm{high}$/Acc$_\mathrm{high}$ and EM$_\mathrm{low}$/Acc$_\mathrm{low}$ are restricted to high- and low-interaction transitions. Fine-tuned models (SFT) are evaluated zero-shot. Few-shot results are in Appendix~\ref{app:l1-full}. Bold marks the best prompted or fine-tuned LLM per column, baselines excluded.}
\label{tab:l1}
\vskip -1em
\end{table*}

\section{Experimental Setup}
\label{sec:setup}

We summarize the world model training and evaluation setup and defer full implementation details to Appendix~\ref{app:setup}. Unless otherwise specified, all experiments follow the default settings below.

\paragraph{Data.}
The training and test sets contain 849k and 39k transitions in Diplomacy, and 687k and 88k in Polymarket.
We fine-tune on 20k samples per world, split evenly between Tasks~J and~S.
To limit evaluation costs, we compare all models on shared test subsets of 1,000 transitions in Diplomacy and 1,333 in Polymarket, unless stated otherwise.

\paragraph{Input and Output.}
Each decision point is written as text with the current observation $O_t$, optionally the previous $K$ transitions of the log as history ($K=0$ unless stated otherwise), and the agent's action, plus the other actors' actions in Task~J. LLMs output only their prediction of the next observation. 
In Diplomacy, LLMs predict the next board.
% In Polymarket, LLMs predict the probability that the mid-price remains unchanged, and three quantiles of its change. 
In Polymarket, LLMs predict future buy and sell prices, the mid-price change, and its uncertainty. 
Prompts are in Appendix~\ref{app:prompts}.

\paragraph{Models.}
We prompt two closed models, GPT-6-Luna and GPT-6-Sol, and fine-tune Qwen3-1.7B and Qwen3-8B~\citep{yang2025qwen3} (Appendix~\ref{app:models}). 
Prompting tests whether pretrained LLMs can predict future observations, while fine-tuning tests whether event logs improve their predictions.

\paragraph{Baselines.}
In Diplomacy, we keep the board unchanged, or apply the given orders literally as if every order succeeded (in Task~S only the agent's own orders). In Polymarket we report a no-change forecast and LightGBM~\citep{ke2017lightgbm}, a gradient-boosted tree model on hand-crafted features of all 687k training points (Appendix~\ref{app:baselines}).

\begin{figure}[t!]
\centering
\includegraphics[width=\columnwidth]{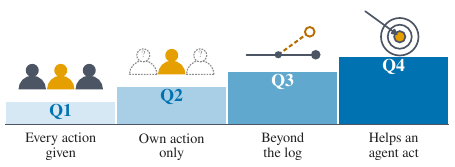}
\caption{The evaluation framework. The agent is orange, actors whose actions are hidden are dashed, and the dashed branch is an action never taken.}
\label{fig:framework}
% \vskip -1em
\end{figure}

\paragraph{Metrics.}
In Diplomacy, exact match (EM) counts a prediction as correct only if the whole next board is right. Since a single wrong unit fails the whole board, F1 also scores units one by one, matching each predicted unit (its player, type, and location) against the true board. In Polymarket, accuracy (Acc) checks whether the model predicts the right direction, that is, whether the mid-price falls, does not change, or rises. 
% The continuous ranked probability score (CRPS)~\citep{gneiting2007strictly} also rewards well-placed uncertainty, measuring how far the predicted distribution of the price change lies from the actual change. 
The continuous ranked probability score (CRPS)~\citep{gneiting2007strictly} evaluates the price-change distribution constructed from the model's outputs (Appendix~\ref{app:metrics}).
We label Diplomacy movement phases and Polymarket hours with at least two orders from other traders as \emph{high-interaction}, and the rest as \emph{low-interaction}.
% Other metrics are introduced where they are used (Appendix~\ref{app:metrics}).

\section{Can LLMs Be Event World Models?}
\label{sec:findings}
\label{sec:framework}
\label{sec:rqs}

% We answer this question with the evaluation framework of Figure~\ref{fig:framework}, whose four questions start from the easiest case and become harder one step at a time. By Eq.~\eqref{eq:tasks}, predicting what an action leads to combines two parts, how the world responds once every action is known and what the other actors will do. \textbf{Q1} tests how well LLMs predict future observations given every actor's action (Task~J). \textbf{Q2} tests the same predictions given only the agent's action (Task~S), so the models also have to anticipate what the other actors do. The drop from Q1 to Q2 measures the cost of not knowing them. A world model is also used in situations not in the event log, so \textbf{Q3} asks whether LLMs can simulate them. \textbf{Q4} asks whether they can help an agent act. Q3 and Q4 are asked both with and without the others' actions, so a weakness found in one question can be traced into the next. We study each across data and model sizes and, where possible, in settings unseen in training (Appendix~\ref{app:framework} lists every experiment).

We evaluate LLMs with four questions (Figure~\ref{fig:framework}).
\textbf{Q1} tests prediction given every actor's action, while \textbf{Q2} provides only the agent's action. 
Their performance gap shows how unknown actions affect prediction.
\textbf{Q3} tests predictions over multiple steps and for actions never taken, and \textbf{Q4} tests whether LLMs as event world models help an agent act.
We compare both input conditions in Q3 and Q4 to examine how this gap affects simulation and agent decisions.
% Appendix~\ref{app:framework} lists all experiments, including data and model scaling and robustness tests.

\subsection{Can LLMs Learn the World's Dynamics Given Every Actor's Action?}
\label{sec:q1}
\label{sec:l1-single}
\label{sec:scale}
\label{sec:robust}

Given the event log and joint action (Task~J), we evaluate predictions of the next board in Diplomacy and the price changes in Polymarket against the recorded outcomes.

\textit{LLMs learn the world's dynamics from the log, building on pretraining.} 
As shown in Table~\ref{tab:l1}, even Qwen3-1.7B, fine-tuned on 20k transitions, reaches 59.7\% EM in Diplomacy, well above the best zero-shot prompted LLM. In Polymarket it reaches 66.1\% Acc, close to LightGBM, which is trained on all 687k transitions. The log alone is not enough either. As shown in Table~\ref{tab:scratch}, the same Qwen3-1.7B architecture trained from random initialization on the same 20k transitions reaches only 32.0\% EM in Diplomacy.

\begin{figure}[t]
\centering
\includegraphics[width=\columnwidth]{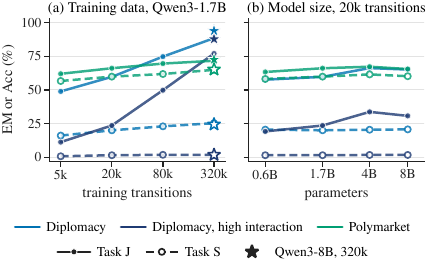}
\caption{Scaling in training data and model size.}
\label{fig:scale}
% \vskip -1em
\end{figure}

\textit{Data matters more than model size.} 
% If the dynamics are learned from the log, more of the log should help. 
As shown in Figure~\ref{fig:scale}, Task~J EM in Diplomacy and Acc in Polymarket rise steadily as the number of training transitions grows. However, larger models improve performance much less than more data does. 
% \lige{Could we clarify what ``given every actor's action'' means for Task J in Polymarket? As I understand it, the order book remains hidden, so unlike in Diplomacy, the next state is not fully determined even when all actors' actions are given. Since Task J still involves uncertainty from the hidden market state, I feel a bit confused about it.}

% \bohan{The appendix clarifies that all three horizons are included in fine-tuning. It would help to mention this alongside Figure 4a to avoid confusion with generalization to unseen horizons.}
\textit{Fine-tuned LLMs learn how much the world changes as time passes.} 
An LLM that has learned this should know that the longer the horizon $\Delta$, the more likely the world has changed. Figure~\ref{fig:l2-clock}a tests this in Polymarket, where the horizon can vary. It compares each model's predicted probability that the price does not change with the observed frequency, the share of test windows in which the price did not change. The Qwen3-1.7B, which saw all three horizons in fine-tuning, matches the observed frequency at each of them. Prompted LLMs overestimate it and the gap grows with the horizon. 

\textit{What fine-tuned LLMs learn transfers to unseen settings of the same world.} 
If the models learned the dynamics rather than the training games and markets, they should stay robust when the setting changes.
On two-player Diplomacy games, the fine-tuned LLMs keep a clear margin over applying the orders literally. On markets that open after the training period, they keep or improve their Acc, while LightGBM falls behind them (Appendix~\ref{app:robust}).

\begin{findingbox}{Finding 1: The world's dynamics}
\textbf{Finding.} Given every actor's action, LLMs can learn an event world's dynamics once fine-tuned on its log.\\
\textbf{Implication.} Building an event world model therefore calls for collecting more event logs and fine-tuning a pretrained LLM on them.
\end{findingbox}

\subsection{Can LLMs Learn the World's Dynamics Given Only the Agent's Own Action?}
\label{sec:q2}
\label{sec:l2}
\label{sec:l2-gap}
\label{sec:l2-history}
\label{sec:l2-clock}

An agent does not know the others' actions in advance. Without them (Task~S), an LLM also has to anticipate the other actors.

% \wei{make clear the difference between an agent and actors}\lige{I feel this is now clear as in Eq~\ref{eq:agent}, where Actor is $A^{-i}$, and agent is $A^i$}

\begin{figure}[t!]
\centering
\includegraphics[width=\columnwidth]{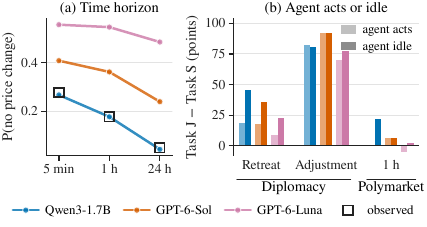}
\caption{(a) Predicted probability that the Polymarket price does not change against the observed frequency. (b) Gain from knowing the other actors' actions (Task~J minus Task~S) when the agent acts and when it is idle.}
\label{fig:l2-clock}
% \vskip -1em
\end{figure}

\textit{Not knowing the other actors' actions hurts most when their actions, rather than the agent's, determine the next observation.} Without their actions, the fine-tuned Qwen3-8B falls from 65.1\% to 20.8\% EM in Diplomacy (Table~\ref{tab:l1}). On high-interaction transitions no model beats applying the agent's own orders and assuming the others do nothing. By contrast, in Polymarket the drop is smaller, because the agent's own large order may drive the price. As shown in Figure~\ref{fig:l2-clock}b, knowing the other traders' orders raises Qwen3-1.7B's Acc substantially when the agent is idle, but barely when its large order drives the price. In Diplomacy retreat phases, the gain roughly doubles when the agent is idle. In adjustment phases, the gain is large either way, because the other players issue most of the orders whether the agent acts or is idle.

\begin{figure}[t]
\centering
\includegraphics[width=\columnwidth]{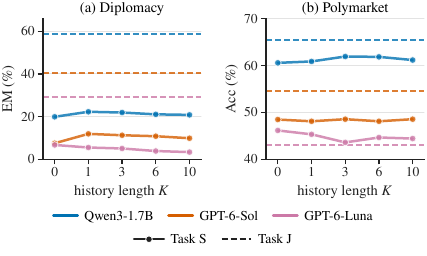}
\caption{Task~S against history length $K$. The full sweep is in Appendix~\ref{app:l2-history}.}
\label{fig:hist}
\vskip -1em
\end{figure}

\textit{The gap remains with history, more data, and larger models.} As shown in Figure~\ref{fig:hist}, history raises Task~S a little for Qwen3-1.7B in both worlds and more for GPT-6-Sol in Diplomacy, while GPT-6-Luna gets worse. 
% Predicting the other actors' actions first lifts the fine-tuned LLMs' Task~S only slightly, from 20.8\% to 22.9\% EM in Diplomacy for Qwen3-8B and from 59.9\% to 60.4\% Acc in Polymarket for Qwen3-1.7B (Table~\ref{tab:l2-gap-full}). It helps GPT-6-Sol more, from 7.5\% to 11.9\% in Diplomacy and from 48.5\% to 52.6\% in Polymarket. 
Asking models to also predict the other actors' behavior improves the fine-tuned models' Task~S scores only slightly, from 20.8\% to 22.9\% EM in Diplomacy for Qwen3-8B and from 59.9\% to 60.4\% Acc in Polymarket for Qwen3-1.7B (Table~\ref{tab:l2-gap-full}). The gains are larger for GPT-6-Sol, from 7.5\% to 11.9\% in Diplomacy and from 48.5\% to 52.6\% in Polymarket.
% As shown in Figure~\ref{fig:scale}, more data raises Task~S only slowly, and a larger model barely changes it. However, for the fine-tuned LLMs and GPT-6-Sol, Task~S with history stays below Task~J. The LLMs thus still anticipate the other actors poorly.
As shown in Figure~\ref{fig:scale} and Table~\ref{tab:scaling}, more data improves Polymarket Task~S overall. In Diplomacy, Task~S improves mainly on low-interaction transitions. Larger models help little in either world. For the fine-tuned LLMs and GPT-6-Sol, Task~S with history also remains below Task~J in both worlds.

\begin{findingbox}{Finding 2: The other actors}
\textbf{Finding.} Given only the agent's action, LLMs learn an event world's dynamics poorly where actions interact.\\
\textbf{Implication.} Modeling the other actors is a problem in its own right. Future work should model uncertainty in their joint choices and examine whether signals about their goals help predict their actions.
\end{findingbox}

\subsection{Can LLMs Simulate Situations Not in the Event Log?}
\label{sec:q3}
\label{sec:l1-rollout}
\label{sec:l3}
% \wei{why we wanna study this problem? show studying this problem is useful}
An agent uses a world model to try out its options before acting. This requires predictions in two situations not in the event log: the observations several steps ahead, where the model builds on its own predictions rather than on recorded observations (a long-horizon rollout), and the outcome of an action the agent never took. 
% A rollout can still be compared with what the log records later, but the effect of an action never taken has to be found outside the log~\citep{oberst2019counterfactual}. In Diplomacy the game engine computes it. Polymarket has no such engine, so we test actions never taken only in Diplomacy (Appendix~\ref{app:l2l3} checks the direction of the effect in Polymarket).
A rollout can be checked against later observations in the log, but logs alone cannot establish the effect of an action never taken~\citep{oberst2019counterfactual}. We use the Diplomacy engine to check predictions for alternative orders. In Polymarket, Appendix~\ref{app:pm-response} compares predicted responses to modified orders with average price responses in the logs.

\textit{LLMs' errors accumulate during a rollout, especially when the other actors' actions are not given.} 
% If small errors stayed small, a rollout would track the log. 
After a five-step rollout in Diplomacy, most models have EM$\approx0$, so we report F1.
As shown in Table~\ref{tab:l1-rollout}, when fed their own predictions in Task~J, the prompted LLMs lose much more F1 than the fine-tuned LLMs. 
Fine-tuning therefore reduces error accumulation. However, in Task~S every model loses much of its F1 as it also has to anticipate the other actors' actions. In Polymarket, the rollout recursively predicts prices while reading trading activity from the log each hour. The fine-tuned LLMs' direct 24-hour forecasts are more accurate than their rollouts even with these activity updates.
% \bohan{Should we clarify that this finding applies to the fine-tuned models in the Polymarket 24-hour forecasting experiment?}

\begin{table}[t]
\centering
\small
\setlength{\tabcolsep}{2.8pt}
\begin{tabular}{@{}llcccc@{}}
\toprule
& & \multicolumn{2}{c}{Diplomacy} & \multicolumn{2}{c}{Polymarket} \\
\cmidrule(lr){3-4}\cmidrule(l){5-6}
Task & Model & F1$_\mathrm{closed}$ & F1$_\mathrm{open}$ & Acc$_\mathrm{closed}$ & Acc$_\mathrm{direct}$ \\
\midrule
\multirow{4}{*}{J} & Qwen3-1.7B & 91.9 & 97.5 & 60.2 & 64.4 \\
 & Qwen3-8B & 94.1 & 97.9 & 61.2 & 63.8 \\
 & GPT-6-Luna & 65.7 & 91.7 & 57.4 & 35.6 \\
 & GPT-6-Sol & 83.4 & 96.4 & 57.4 & 55.6 \\
\midrule
\multirow{4}{*}{S} & Qwen3-1.7B & 55.1 & 86.0 & 55.6 & 58.6 \\
 & Qwen3-8B & 55.4 & 86.5 & 57.0 & 59.0 \\
 & GPT-6-Luna & 52.8 & 85.9 & 56.8 & 43.2 \\
 & GPT-6-Sol & 53.9 & 86.6 & 54.0 & 53.4 \\
\bottomrule
\end{tabular}
\caption{Long-horizon rollout results. Diplomacy: F1 at step 5 when the model is fed its own predictions (closed loop) or the true board (open loop). Polymarket: Acc over 24 hours from recursive price prediction with logged trading activity (closed) or a direct forecast. Full table in Appendix~\ref{app:rollout}.}
\label{tab:l1-rollout}
% \vskip -1em
\end{table}

\textit{Without the other actors' actions, LLMs misjudge the effect of an action never taken. Given those actions, the fine-tuned LLMs misjudge it far less.} An LLM that has learned the world's dynamics should still predict the next observation accurately when the agent's action is changed. However, where the new orders do not change the next board, the fine-tuned LLMs still predict a change in more than half of the cases in Task~S (spurious, Table~\ref{tab:l3}). Giving the other players' recorded orders should then remove much of the error from imagining their orders wrongly, or differently for the agent's original and new orders. It sharply raises the fine-tuned LLMs' F1 and sharply lowers their spurious rate. The prompted GPT-6 models, in contrast, do not improve, and they predict a change in most of the cases where there is none. The log teaches how the world responds to a fixed joint action. When the model has to imagine the others, its prediction for an action never taken therefore mixes the effect of that action with the effect of wrong or inconsistent imagined orders.

\begin{table}[t]
\centering
\small
\setlength{\tabcolsep}{6pt}
\begin{tabular}{@{}lcccc@{}}
\toprule
& \multicolumn{2}{c}{F1} & \multicolumn{2}{c}{Spurious$\downarrow$} \\
\cmidrule(lr){2-3}\cmidrule(l){4-5}
Model & S & J & S & J \\
\midrule
GPT-6-Luna & 47.7 & 30.1 & 45.9 & 89.7 \\
GPT-6-Sol & 56.5 & 57.9 & 40.4 & 83.3 \\
Qwen3-1.7B & 44.0 & 73.0 & 51.5 & 33.1 \\
Qwen3-8B & 44.3 & 77.3 & 59.6 & 29.4 \\
\bottomrule
\end{tabular}
\caption{Results after replacing the agent's orders in Diplomacy (\%).
F1 compares predicted and true changes to units and supply centers.
Spurious is the rate of predicting a change when original and modified orders produce the same board.
More detailed results are reported in Appendix~\ref{app:l2l3}.}
\label{tab:l3}
% \vskip -1em
\end{table}

\begin{findingbox}{Finding 3: Beyond the log}
\textbf{Finding.} LLMs simulate situations not in the event log poorly. Their rollouts accumulate errors, and they misjudge the effects of actions never taken, the fine-tuned ones far less so once the other actors' actions are given.\\
\textbf{Implication.} Event world models should account for uncertainty about other actors when comparing alternative actions. They should also use new observations to correct prediction errors accumulated during rollouts.
\end{findingbox}
% \wei{making naming consistent; should we say "event world model"}

\subsection{Can LLMs as Event World Models Help an Agent Act?}
\label{sec:q4}
\label{sec:agent}

So far we have evaluated LLMs as world models on their own. We further ask whether LLMs can serve an agent and help it perform better. Answering this requires an engine that executes the agent's actions, so we use Diplomacy. The agent plays against six other actors controlled by DipNet~\citep{paquette2019no}, a deterministic policy trained to imitate human players. 
The agent succeeds if it ends 10 phases with at least its initial number of supply centers.

\begin{figure}[t]
\centering
\includegraphics[width=\columnwidth]{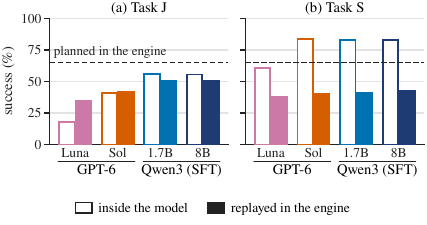}
\caption{Success of plans made inside an LLM, judged inside the LLM and when replayed in the engine, against plans made in the engine.}
\label{fig:plan}
% \vskip -1em
\end{figure}

\textit{Without the other actors' actions, plans look more successful inside an LLM than they turn out to be in the engine.} An LLM can serve an agent as a simulator in which the agent tries out its plans before acting. If an LLM captured how the opponents play, plans made inside it would work as well in the engine. The agent plays each game in the engine and inside the LLM, and its orders from inside the LLM are replayed in the engine. As shown in Figure~\ref{fig:plan}b, in Task~S the agent succeeds more often inside the fine-tuned LLMs and GPT-6-Sol than in the engine. However, replayed in the engine, these plans succeed far less often than plans made in the engine. Even in Task~J, replayed plans succeed less often (Figure~\ref{fig:plan}a).

\begin{figure}[t]
\centering
\includegraphics[width=\columnwidth]{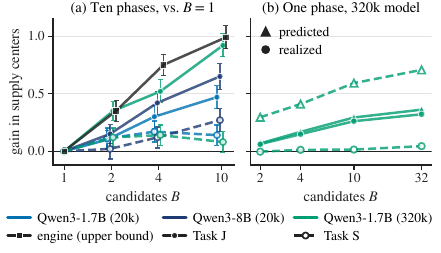}
\caption{Evaluation of candidate actions by LLMs in Diplomacy. (a) Gain in supply centers over $B{=}1$. (b) Predicted and realized gain of the chosen candidate.}
\label{fig:agent-verifier}
% \vskip -1em
\end{figure}

\textit{Choosing among candidate actions with an LLM helps mainly when the joint action is known.} 
% Many actions cannot be undone, so one single mistake can be costly. An LLM can then serve the agent by checking its options before it acts. 
At every movement phase, the agent samples $B$ candidate actions, the LLM predicts the board after each, and the agent chooses the best one. As shown in Figure~\ref{fig:agent-verifier}a, in Task~J the agent gains more supply centers as more candidates are offered.  With $B{=}10$ candidates, Qwen3-1.7B trained on 320k transitions nearly matches the engine. However, in Task~S the agent gains almost nothing with the same LLM. The LLM's predicted gain grows with more candidates while the realized gain does not, so it overrates its pick (Figure~\ref{fig:agent-verifier}b).

\begin{findingbox}{Finding 4: Decisions}
\textbf{Finding.} LLMs help an agent choose its actions mainly when the other actors' actions are given. Otherwise, plans look better inside an LLM than in reality, and the agent picks actions that the LLM overrates.\\
\textbf{Implication.} When planning, an agent should account for a distribution over the other actors' actions, rather than a single deterministic prediction of them.
\end{findingbox}

\begin{table}[t]
\centering
\footnotesize
\begin{tabular}{@{}lcc@{}}
\toprule
Training data (Qwen3-1.7B policy) & $\Delta$SC & Success (\%) \\
\midrule
Behavior policy (DipNet) & 0.24 & 61.0 \\
\midrule
Real 1k & $-$0.58 & 47.5 \\
Imagined 1k & $-$0.62 & 43.0 \\
Real 1k + imagined 1k & $-$0.29 & 53.5 \\
Real 2k & $-$0.23 & 54.0 \\
Imagined 2k & $-$0.48 & 47.5 \\
\midrule
Real 1k, init.\ world model (20k) & $-$0.36 & 55.0 \\
Real 1k, init.\ world model (320k) & $-$0.04 & 59.5 \\
\bottomrule
\end{tabular}
\caption{Policies behavior-cloned on real or imagined games, or initialized from a world model. $\Delta$SC: mean change in supply centers over 10 phases.}
\label{tab:agent-bc}
% \vskip -1em
\end{table}

\textit{Imagined games can replace part of the real games, but not all of them.} 
% When real interaction is costly, LLMs can serve the agent by imagining games that replace part of the real ones. 
We train a Qwen3-1.7B policy by behavior cloning on the agent's orders in successful games, played in the engine or inside the 320k LLM given only the agent's orders. As shown in Table~\ref{tab:agent-bc}, 1k imagined games train a policy nearly as good as 1k real games. Adding 1k imagined games to 1k real ones also helps nearly as much as 1k more real games. However, 2k imagined games train a weaker policy than 2k real games, so more imagined games add less than more real ones (Appendix~\ref{app:agent}).
% \lige{Have we evaluated if policy performance scales with the number of imagined games? It might be interesting to see whether generating more imagined experience continues to improve policy learning (or maybe not).}

\textit{Starting the policy from a world model helps, and more so from a better one.} Recent work~\citep{zhang2025agent} suggests that exposing a model to environment dynamics before policy learning gives it a useful inductive bias. A world model can thus also serve as the starting point of the agent's policy. As shown in Table~\ref{tab:agent-bc}, a policy cloned from 1k real games does better when initialized from the 20k world model, and better still from the 320k one.

% \bohan{Should we discuss the reading control in Appendix M here, including its implications and limitations for attributing the initialization gains to dynamics learning?}

\begin{findingbox}{Finding 5: Learning}
\textbf{Finding.} LLMs also help an agent learn, since imagined games can replace part of the real ones. Initializing the policy from an event world model helps too.\\
\textbf{Implication.} Without a simulator, an event world model learned from logs could potentially be both the starting point and part of the training of an agent, with scarce real interaction spent on the rest.
\end{findingbox}

\section{Conclusion}
\label{sec:conclusion}
We define event worlds and introduce two complementary benchmarks, Diplomacy and Polymarket. Our evaluation covers prediction, simulation beyond the log, and usefulness to agents. Fine-tuned LLMs learn world responses given all actors' actions, but struggle when only the agent's action is known. More data and larger models do not close this gap. In Diplomacy, this limitation also affects predictions for alternative actions and agent decisions. These findings highlight modeling other actors' unknown actions as a central direction for improving event world models.

\clearpage

\section*{Limitations}
Our study has two main limitations. First, it covers only Diplomacy and Polymarket, and the models receive observations prepared from event logs. Broader domains and evaluation directly from raw event streams are needed to assess how well the findings generalize. Second, we verify the outcomes of alternative actions and evaluate agents only in Diplomacy, using its rules engine. The agent experiments use fixed opponent policies, so their findings may not extend to interactions with humans or opponents that learn over time. The Polymarket intervention analysis relies on associations in observed trades and does not establish causal effects.

\section*{Potential Risks}
Our experiments show that LLMs can overestimate plan success when other actors' actions are unknown. Relying on these predictions without further checks could lead to poor decisions, especially if the models are applied to financial or medical settings. Applications should be evaluated using only information available before acting and should make prediction uncertainty clear.

\bibliography{acl}

\clearpage
\appendix

\begin{table*}[ht!]
\centering
\small
\setlength{\tabcolsep}{4pt}
\begin{tabularx}{\textwidth}{@{}>{\raggedright\arraybackslash}p{3.0cm}>{\raggedright\arraybackslash}X>{\raggedright\arraybackslash}X@{}}
\toprule
Line of work & Main prediction or evaluation target & Relation to our study \\
\midrule
Text, web, tool world models & Observations after an agent's actions & We extend evaluation to logs with multiple actors whose future actions are unknown. \\
Visual and driving world models & Frames, scene states, or joint trajectories & Event logs can record actions and observations at irregular times. \\
Multi-agent benchmarks & Agents' collaboration and competition & We evaluate world predictions from human game records and real market logs, then their usefulness to an agent. \\
Models of other actors & Other participants' actions or beliefs & These predictions are one part of predicting the world's future observations. \\
Event forecasting & Future events from records or news & Our predictions additionally condition on the agent's proposed actions. \\
Offline model-based RL & Dynamics and policies learned from logged transitions & We examine the difficulty introduced by unknown actions of other actors. \\
Market and clinical simulators & Outcomes following trades or treatments & These are close domain-specific studies. We provide a unified formulation and evaluation across two worlds. \\
\bottomrule
\end{tabularx}
\caption{Related work compared by its prediction or evaluation target and its relation to event world modeling. Representative studies are discussed in Appendix~\ref{app:related}.}
\label{tab:related}
\end{table*}

\section{Extended Related Work}
\label{app:related}

Table~\ref{tab:related} compares the lines of work closest to event worlds. \citet{chu2026agentic} sort world models into physical, digital, social, and scientific regimes by the laws that govern each world. Event worlds span these regimes, defined by how they change, who acts in them, and how they are observed.

\paragraph{LLM World Models of Text, Web, and Tool Environments.}
LLMs predict the next state of text, web, and tool environments~\citep{li2026word,zuo2026qwen} and help agents plan or train~\citep{gu2024your,li2026word}. These studies usually evaluate interaction in agent steps, with an observation returned after each action, including when training uses real software or the open web~\citep{zuo2026qwen,xiao2026webworldlargescaleworldmodel}. \citet{wang2024can} also model changes made by the environment between agent steps and find GPT-4 much less accurate on them. Our evaluation framework (Section~\ref{sec:framework}) extends that of \citet{li2026word} to prediction from multi-actor event logs, where other actors' future actions are unknown to the agent.

\paragraph{World Models of Video, Games, Robots, and Driving.}
Visual world models predict dense observations, such as frames, from the agent's actions~\citep{ha2018world,hafner2025mastering,yang2024learning}. Some learn from recorded video instead of a simulator~\citep{bruce2024genie,hafner2025training}. Learned models also support planning in board games such as Go and chess, where two players take turns and training uses self-play~\citep{schrittwieser2020mastering}. Driving models add other road users and an environment that changes without the agent's action~\citep{hu2023gaia}. Traffic simulators learn the motion of many road users from real logs and are evaluated on the realism of joint trajectories~\citep{montali2023waymo,wu2024smart}. Visual and driving models typically predict frames or trajectories sampled at regular intervals. Event logs can instead record actions and observations at irregular times, and our models take the prediction horizon $\Delta$ as an input.

\paragraph{Agent Societies, Game Agents, and Models of Other Agents.}
In agent societies, LLM agents act in designed sandboxes and are judged by how believable their behavior looks~\citep{park2023generative,vezhnevets2023generativeagentbasedmodelingactions}. Diplomacy agents learn from game records how humans play and use the rules engine to resolve their actions~\citep{paquette2019no,meta2022human,bakhtin2022mastering}. These studies model the actors' behavior within a given world. We evaluate predictions of the world's future observations and how these predictions help an agent.

Models of other agents, from opponent models to social world models, predict what others will do or believe~\citep{albrecht2018autonomous,zhou2026socialworldmodels}. An event world model needs such predictions as one part of its task, the first factor of Eq.~\eqref{eq:tasks}. A sequence model trained only on transcripts of game moves forms an internal representation of the board~\citep{li2022emergent}, and \citet{vafa2024evaluating} test whether such models hold a coherent world model. These works ask what a sequence model knows about the state, whereas an event world model predicts the next observation that given actions lead to.

\paragraph{Multi-Agent Benchmarks.}
Multi-agent benchmarks evaluate how agents cooperate or compete in shared environments~\citep{guo2024large,zhu2025multiagentbench}. Many generate interactions using LLM agents in predefined environments. Our prediction tasks use human game records and real market logs, and evaluate a model's predictions of the environment. We then test whether those predictions improve an agent's decisions or learning.

\paragraph{Event Forecasting.}
Neural point processes and LLM-based forecasters predict what comes next in logs of typed events~\citep{mei2017neural,pmlr-v119-zuo20a,shi2023language,lee2023temporal}. LLMs also forecast how real-world questions resolve~\citep{halawi2024approaching,ICLR2025_ea74e45a}. These tasks ask what will happen, not what will happen if a given actor acts. On Polymarket, LLMs forecast outcomes and trade under simulated execution~\citep{cheng2026polybenchbenchmarkingllmforecasting}. \citet{yu2026building} build a social world model in which an LLM predicts how prediction-market prices change after news events. In both, the other actors appear only in aggregate, through prices or the order book, and neither learns how the market responds to an order.

\paragraph{Learning Decisions from Logs.}
Offline reinforcement learning (RL) learns from logged data without further interaction~\citep{levine2020offlinereinforcementlearningtutorial}. Its model-based methods learn how actions change a fixed-length state vector~\citep{yu2020mopo,NEURIPS2020_f7efa4f8}, and sequence models of logged trajectories can serve the same role~\citep{janner2021offline}. However, without further assumptions, logs alone cannot determine what would have happened under an action never taken~\citep{oberst2019counterfactual}.

Simulators learned from the logs of real systems come closest, for markets~\citep{coletta2022learning,li2025mars}, online retail~\citep{shi2019virtual}, and hospital stays~\citep{mu2026ehrworld}. EHRWorld fine-tunes an LLM on hospital records to predict a patient's next state given the treatments. We build on these domain-specific studies with a unified formulation and evaluation across two event worlds. Comparing prediction with and without the other actors' recorded actions shows how their unknown choices affect prediction and usefulness to an agent.

\section{From Raw Records to Event Worlds}
\label{app:bench}

Table~\ref{tab:instantiation} shows how each world instantiates the definition of Section~\ref{sec:ewm-def}.

\begin{table}[h]
\centering
\footnotesize
\setlength{\tabcolsep}{3pt}
\begin{tabularx}{\columnwidth}{@{}l>{\raggedright\arraybackslash}X>{\raggedright\arraybackslash}X@{}}
\toprule
& Diplomacy & Polymarket \\
\midrule
clock & phase counter & physical time in seconds \\
horizon $\Delta$ & one phase & chosen; 1\,h (also 5\,min, 24\,h) \\
actors $\mathcal{I}$ & 7 (or 2) players & taker wallets \\
action $A^i_{t+1}$ & one order per unit; hold if none & 0 to many orders (time, direction, size) \\
log $\mathcal{H}_{t}$ & boards, orders, results & orders in earlier blocks \\
observation $O_t$ & units, centers & prices, flow, depth proxies \\
hidden $X_t$ & players' intentions & order book, resting quotes, beliefs \\
reference for changed actions & engine outcomes & average logged price responses \\
\bottomrule
\end{tabularx}
\caption{How the general definition maps to the two worlds. Changed actions have engine-verified outcomes in Diplomacy. Polymarket provides an observational response check (Appendix~\ref{app:pm-response}).}
\label{tab:instantiation}
\end{table}

\subsection{Diplomacy}
\label{app:dip}

\paragraph{Observation.}
The observation at a phase, the board, consists of the phase type, every player's units (army or fleet, each at a location), and every player's supply centers. A retreat phase adds which units were dislodged and where they may retreat. An adjustment phase adds how many units each player may build or has to disband. The whole board is visible to every player.

\paragraph{Actions and the Engine.}
In a movement phase a unit can hold, move (possibly by convoy), support, or convoy. Retreat phases allow retreat and disband orders, and adjustment phases allow build and disband orders. The engine resolves the joint action deterministically and returns a result for every unit, such as bounce or dislodged.

\paragraph{Game Records.}
The main experiments use only no-press games, without messages between players. We keep a game only if the engine, replaying its recorded orders, reproduces every recorded board, so that the engine is the ground truth for any joint action. This keeps 33,194 of the 33,279 no-press games (99.7\%, 1.18M transitions). Of these, 21,832 are seven-player games (940k transitions). The other 11,362 are two-player games (237k transitions) of France against Austria or Germany against Italy, played on the standard map.

\paragraph{Decision Points.}
Each transition yields two Task~S samples, one for each of two players drawn at random among those that still have units. The model receives neither the map nor legal orders.

\subsection{Polymarket}
\label{app:pm}

\paragraph{Order Stream and Observation.}
From the public on-chain records of \citet{wang2026polymarket}, we keep the 4,176 markets traded for at least 30 days with at least 5,000 trades. They hold 79M orders from December 2022 to July 2026. Each market is a binary prediction market with static context (question text, parent event, topic, opening time, announced end date). Its record is the stream of taker orders, the orders that trade at once against a resting quote. An order aggregates all fills of the same taker, block, outcome token, and direction. It has a block number, a timestamp, the taker wallet, a direction (buy or sell of \textsc{Yes}), a size in USD, and the volume-weighted execution price.
For a window starting at $t$, the history contains every order in a block before the first block at or after $t$. The observation summarizes this history. The mid-price proxy is the mean of the latest available buy and sell execution prices. Their difference gives a spread proxy. The USD volume of the previous 24 hours is a depth proxy. The number, signed net flow, and volume of the previous hour's orders describe recent activity. The market's age, the time to its announced end, and the static context complete the observation. The input also gives the agent's net position, past order count, and past flow, derived from the history.

\paragraph{Actors and the Next Observation.}
Actors in this benchmark are taker wallets. Makers' resting quotes are unobserved and enter through the hidden order book. The prediction task asks for future buy and sell prices and the change in their midpoint. Other fields of later observations are computed from the recorded log. Future prices remain uncertain even when all recorded orders are given, because they also depend on the hidden \mbox{order book (Section~\ref{sec:worlds})}.

\paragraph{Decision Points.}
An anchored window starts at a large order, one whose relative size (its USD size divided by the volume of the previous 24 hours) is at least 0.1 and whose wallet placed no order in the previous hour. This wallet is the agent. Control windows are drawn uniformly in time and have no designated agent, so Task~S receives no action there, as for an idle agent. Every decision point lies at least 24 hours after the market opened, and its window ends before the market's last trade. Windows last one hour by default, the shortest horizon at which other traders have usually responded.
Task~J gives every order in the window with its time, wallet, direction, and size, but not its execution price, and asks for the prices at the end of the window. Task~S gives only the agent's orders. It gives all of them in the window, not only the first, because 45\% of large orders are followed by further orders of the same wallet within the hour.

\begin{table*}[t]
\centering
\small
\setlength{\tabcolsep}{4pt}
\renewcommand{\arraystretch}{1.1}
\begin{tabularx}{\textwidth}{@{}>{\raggedright\arraybackslash}p{2.0cm}>{\raggedright\arraybackslash}X>{\raggedright\arraybackslash}p{2.7cm}>{\raggedright\arraybackslash}X>{\raggedright\arraybackslash}X>{\raggedright\arraybackslash}p{0.75cm}@{}}
\toprule
Question & Experiment & Evaluation reference & Diplomacy & Polymarket & Sec. \\
\midrule
Q1: every action given & Does the model predict the next observation when every actor's action is given? & the log & next board from all players' orders & price move over the window from all orders & \ref{sec:q1} \\
 & Does more data or a larger model help, and is pretraining needed? & the log & 5k to 320k transitions; 0.6B to 8B parameters; training from scratch & same & \ref{sec:q1} \\
 & Does the model know how much the world changes as time passes? & the log & & horizons of 5\,min, 1\,h, and 24\,h & \ref{sec:q1} \\
 & Does what the model learns hold when the world shifts? & the log & two-player games, unseen in training & markets that open after the training period & \ref{sec:q1} \\
\midrule
Q2: only the agent's action & How much does knowing the other actors' actions improve prediction? & the log & Task~J vs.\ Task~S, by interaction and when the agent issues no orders & Task~J vs.\ Task~S, including windows where the agent does nothing & \ref{sec:q2} \\
 & Do history, predicting the other actors' behavior, or scale close the gap? & the log & up to 10 previous phases; factored variant, its orders resolved by the engine; data and model size & up to 10 previous hours; factored variant & \ref{sec:q2} \\
\midrule
Q3: beyond the log & Does accuracy survive when the model builds on its own predictions? & the log & five-step chains & 24 hourly price updates with logged trading activity & \ref{sec:q3} \\
& How do predictions respond to changed actions? & engine outcomes; average logged responses & the agent's orders replaced, the other players' orders hidden or given & the agent's orders modified for an observational response check & \ref{sec:q3} \\
\midrule
Q4: helping an agent act & Do plans made inside the model work in the engine? & engine outcomes & plans made in the model replayed in the engine & not evaluated & \ref{sec:q4} \\
& Does the model help an agent choose actions and learn? & engine outcomes & choosing among candidate actions; imagined games; policy initialization & not evaluated & \ref{sec:q4} \\
\bottomrule
\end{tabularx}
\caption{The evaluation framework of Section~\ref{sec:framework}, one experiment per row.}
\label{tab:framework}
\end{table*}

\subsection{Splits and Evaluation Sets}
\label{app:data}

\paragraph{Diplomacy.}
We sort the seven-player games by identifier. The last 4\% form the test split, aligned with the held-out tail of the DipNet training data~\citep{paquette2019no}. A buffer of 330 games before the test split goes to validation, so that no game seen by DipNet enters the test split. The remaining games are divided 95/5 into training and validation by a hash of the game identifier. This gives 19,721 training games (849k transitions), 1,212 validation games, and 899 test games (39,224 transitions). The validation games serve only to set the learning rate of the from-scratch control (Appendix~\ref{app:scratch}). From the test split we draw a \emph{common evaluation set} of 3,000 transitions (1,471 movement, 860 retreat, 669 adjustment). The \emph{shared subset}, drawn from it, has 1,000 transitions (504 movement, 282 retreat, 214 adjustment) and is used in the main text unless stated otherwise. The two-player common evaluation set has 1,000 transitions (581 movement, 181 retreat, 238 adjustment). No two-player game is used for training.

\paragraph{Polymarket.}
All markets of one parent event fall in the same split. Events first traded from January 1, 2026, form the later-market split (890 markets). Events first traded from October to December 2025 form a validation split (717 markets), which is not used for any reported result. The remaining events are divided 90/10 by a hash of the event into the training split (2,170 markets) and the in-distribution test split (258 markets). Outside the later-market split, we drop decision points whose 24-hour target window reaches into 2026. This leaves 1.02M of the 1.37M decision points. The market counts above include only markets that keep a decision point. The training split then has 687,464 decision points, and the in-distribution test split has 88k. Each test split has a common evaluation set of 4,000 decision points at $\Delta = 1$\,h, of which 1,000 are control windows. The shared subset of the in-distribution test split has 1,333 points (1,000 anchored and 333 control windows), of which 598 are high-interaction and 735 low-interaction. The later-market split also has a 1,333-point subset, used in Appendices~\ref{app:scratch} and~\ref{app:leak}.

\subsection{Artifacts, Licenses, and Privacy}
\label{app:artifacts}
We use the Diplomacy game records released by \citet{paquette2019no}, which were played on webDiplomacy and are provided for research use. We also use the on-chain Polymarket trade records of \citet{wang2026polymarket}, released under the MIT license. We fine-tune Qwen3~\citep{yang2025qwen3}, released under the Apache~2.0 license, and query GPT-6-Luna and GPT-6-Sol through their provider's API under its terms of use. Other software includes the open-source Diplomacy engine (AGPLv3), the DipNet policy~\citep{paquette2019no}, LLaMA-Factory~\citep{zheng2024llamafactory} and vLLM~\citep{kwon2023efficient} (both Apache~2.0), and LightGBM~\citep{ke2017lightgbm} (MIT). We use all of them for non-commercial research, consistent with their terms, and the benchmarks derived from them are intended for research on event world models. The Diplomacy records contain no player names, since each player is identified only by the country they play, and the main experiments use games without messages. The negotiation experiment in Appendix~\ref{app:negotiation} uses the in-game messages as released, without filtering. Polymarket wallets are pseudonymous public blockchain addresses, and every prompt replaces them with per-sample pseudonyms (the agent is A, and the other traders are w01, w02, and so on in order of appearance).

% \FloatBarrier
\section{Evaluation Framework in Detail}
\label{app:framework}

Table~\ref{tab:framework} lists the experiments of Section~\ref{sec:findings}.

\FloatBarrier
\section{Training and Evaluation Details}
\label{app:setup}

\subsection{Prompts and Serialization}
\label{app:prompts}

\paragraph{System Prompts.}
Each request has a system prompt and a user turn with the serialized decision point. Few-shot requests add three demonstrations from the training split before the user turn, each an input with its recorded next observation. In Diplomacy they are fixed transitions, one per phase type. The Polymarket listings below show the prompted models' format. Fine-tuned Polymarket models use the format described in Appendix~\ref{app:pm-output}.

\begin{lstlisting}[basicstyle=\ttfamily\scriptsize,breaklines=true,frame=single,columns=fullflexible,title={System prompt, Diplomacy, Task J}]
You are a world model for Diplomacy. Given the current board (STATE), optionally a HISTORY of previous phases, and every player's orders for this phase (ORDERS), output a line 'NEXT <phase>' and the STATE block of the board after this phase is resolved, in the same format as the input. Output nothing else.
\end{lstlisting}

\begin{lstlisting}[basicstyle=\ttfamily\scriptsize,breaklines=true,frame=single,columns=fullflexible,title={System prompt, Diplomacy, Task S}]
You are a world model for Diplomacy. Given the current board (STATE), optionally a HISTORY of previous phases, and one player's orders for this phase (OWN), output a line 'NEXT <phase>' and the STATE block of the board after this phase is resolved, in the same format as the input. Output nothing else.
\end{lstlisting}

\begin{lstlisting}[basicstyle=\ttfamily\scriptsize,breaklines=true,frame=single,columns=fullflexible,title={System prompt, Polymarket, Task J}]
You are a world model for a prediction market. Given the market state and all orders in the window, output a JSON object with keys p_buy, p_sell, p_zero, q10, q50, q90 and nothing else. p_buy and p_sell are PRICE LEVELS in [0,1] at the end of the horizon. q10, q50, q90 are quantiles of the SIGNED CHANGE in mid price from now to the end of the horizon, in price units (for example -0.012 means the mid falls by 1.2 cents; 0.0 means unchanged); they are NOT price levels and must satisfy q10 <= q50 <= q90. p_zero is the probability that the mid is exactly unchanged.
\end{lstlisting}

\begin{lstlisting}[basicstyle=\ttfamily\scriptsize,breaklines=true,frame=single,columns=fullflexible,title={System prompt, Polymarket, Task S}]
You are a world model for a prediction market. Given the market state and the actor's own orders in the window, output a JSON object with keys p_buy, p_sell, p_zero, q10, q50, q90, and nothing else. p_buy and p_sell are PRICE LEVELS in [0,1] at the end of the horizon. q10, q50, q90 are quantiles of the SIGNED CHANGE in mid price from now to the end of the horizon, in price units (for example -0.012 means the mid falls by 1.2 cents; 0.0 means unchanged); they are NOT price levels and must satisfy q10 <= q50 <= q90. p_zero is the probability that the mid is exactly unchanged.
\end{lstlisting}

\paragraph{Example Inputs.}
The examples below use $K=0$. The Diplomacy examples share a board and target. Task~S gives only Russia's empty action, while Task~J gives all players' orders. A and F denote armies and fleets, and SC lists supply centers. With $K=3$, the input adds the previous three phases' boards, orders, and results. In Polymarket, $K=3$ adds three hourly price and flow bars and up to 20 recent orders (Appendix~\ref{app:l2-history}).

\begin{lstlisting}[basicstyle=\ttfamily\scriptsize,breaklines=true,frame=single,columns=fullflexible,title={Diplomacy, Task J, input}]
GAME standard 7P RULES NO_CHECK
PHASE W1905A
STATE
ENG: A WAL F LON F NAO | SC EDI LON
FRA: A BEL A BUR A MAR A PIC F ENG F LVP | SC BEL BRE LVP MAR PAR POR SPA
GER: A BER A DEN A MUN F BAL F NTH | SC BER DEN HOL KIE MUN
ITA: A ALB A SER A TYR F AEG F GRE F ION | SC GRE NAP ROM SER TRI TUN VEN
RUS: A MOS A SWE A WAR F BOT F SKA | SC MOS NWY STP SWE WAR
TUR: A ARM A BUD A GAL A RUM A VIE F BUL/SC F CON F SEV F SMY | SC ANK BUD BUL CON RUM SEV SMY VIE
BUILDS ENG -1 ; FRA +1 @BRE,PAR ; ITA +1 @NAP,ROM,VEN ; TUR -1
ORDERS
ENG: A WAL D
FRA: F BRE B
ITA: A VEN B
TUR: F SEV D
\end{lstlisting}

\begin{lstlisting}[basicstyle=\ttfamily\scriptsize,breaklines=true,frame=single,columns=fullflexible,title={Diplomacy, Task J, reference output}]
NEXT S1906M
STATE
ENG: F LON F NAO | SC EDI LON
FRA: A BEL A BUR A MAR A PIC F BRE F ENG F LVP | SC BEL BRE LVP MAR PAR POR SPA
GER: A BER A DEN A MUN F BAL F NTH | SC BER DEN HOL KIE MUN
ITA: A ALB A SER A TYR A VEN F AEG F GRE F ION | SC GRE NAP ROM SER TRI TUN VEN
RUS: A MOS A SWE A WAR F BOT F SKA | SC MOS NWY STP SWE WAR
TUR: A ARM A BUD A GAL A RUM A VIE F BUL/SC F CON F SMY | SC ANK BUD BUL CON RUM SEV SMY VIE
\end{lstlisting}

\begin{lstlisting}[basicstyle=\ttfamily\scriptsize,breaklines=true,frame=single,columns=fullflexible,title={Diplomacy, Task S, input}]
GAME standard 7P RULES NO_CHECK
PHASE W1905A
STATE
ENG: A WAL F LON F NAO | SC EDI LON
FRA: A BEL A BUR A MAR A PIC F ENG F LVP | SC BEL BRE LVP MAR PAR POR SPA
GER: A BER A DEN A MUN F BAL F NTH | SC BER DEN HOL KIE MUN
ITA: A ALB A SER A TYR F AEG F GRE F ION | SC GRE NAP ROM SER TRI TUN VEN
RUS: A MOS A SWE A WAR F BOT F SKA | SC MOS NWY STP SWE WAR
TUR: A ARM A BUD A GAL A RUM A VIE F BUL/SC F CON F SEV F SMY | SC ANK BUD BUL CON RUM SEV SMY VIE
BUILDS ENG -1 ; FRA +1 @BRE,PAR ; ITA +1 @NAP,ROM,VEN ; TUR -1
OWN RUS: none
\end{lstlisting}

\begin{lstlisting}[basicstyle=\ttfamily\scriptsize,breaklines=true,frame=single,columns=fullflexible,title={Polymarket, Task J, input}]
MARKET 253258 | topic other | now 2023-12-03 05:27:54 UTC | horizon 1h
Q: OpenSea token >1 billion a week after launch?
EVENT: OpenSea token >1 billion a week after launch?
STATE p_buy 0.6300 p_sell 0.5900 mid 0.6100 spread 0.0400 | D24h $3751.66 | last1h N 0 F +$0.00 V $0.00 | age 1.1d ttl 393.8d | last order 12946s ago
AGENT A | position -6085.00 YES | 7 prior orders | net flow -$3573.93
JOINT 2 orders in window (dt wallet side usd): 0s A SELL $789.60 | 2248s w01 BUY $195.00
TASK J: all orders in the window are given above (execution prices hidden). Predict the market state after 1 hour.
OUTPUT JSON {"p_buy": <last taker buy price at t+H>, "p_sell": <last taker sell price at t+H>, "p_zero": <P(mid unchanged)>, "q10": <10% quantile of mid change>, "q50": <median>, "q90": <90% quantile>}
\end{lstlisting}

\begin{lstlisting}[basicstyle=\ttfamily\scriptsize,breaklines=true,frame=single,columns=fullflexible,title={Polymarket, Task S, input}]
MARKET 253258 | topic other | now 2023-12-03 05:27:54 UTC | horizon 1h
Q: OpenSea token >1 billion a week after launch?
EVENT: OpenSea token >1 billion a week after launch?
STATE p_buy 0.6300 p_sell 0.5900 mid 0.6100 spread 0.0400 | D24h $3751.66 | last1h N 0 F +$0.00 V $0.00 | age 1.1d ttl 393.8d | last order 12946s ago
AGENT A | position -6085.00 YES | 7 prior orders | net flow -$3573.93
ACTION A (1 orders, dt side usd): 0s SELL $789.60
TASK S: the agent's orders above are its only orders in the window. Predict the market state after 1 hour.
OUTPUT JSON {"p_buy": <last taker buy price at t+H>, "p_sell": <last taker sell price at t+H>, "p_zero": <P(mid unchanged)>, "q10": <10% quantile of mid change>, "q50": <median>, "q90": <90% quantile>}
\end{lstlisting}

\subsubsection{Fine-Tuned Polymarket Outputs}
\label{app:pm-output}
Each training record supplies a realized price change and the two future execution prices. We encode that change as both a bin label and a numeric value, giving the following output format. The factored variant (SF) adds the other traders' order count, net flow, and volume before these price fields (Appendix~\ref{app:actors}).

\begin{lstlisting}[basicstyle=\ttfamily\scriptsize,breaklines=true,frame=single,columns=fullflexible,title={Fine-tuned Polymarket output format}]
BIN <letter A-U>
DMID <signed mid-price change>
P_BUY <future buy price>
P_SELL <future sell price>
\end{lstlisting}

The 21 bins comprise ten negative-change bins, an exact-zero bin (K), and ten positive-change bins. The nonzero bins use absolute-change boundaries of 0, 0.001, 0.0025, 0.005, 0.01, 0.02, 0.035, 0.05, 0.1, 0.2, and 0.5. At inference, we normalize the probabilities of the 21 bin letters. The zero-bin probability gives \texttt{p\_zero}, and a uniform density within each nonzero bin gives the 10th and 90th percentiles. The generated \texttt{DMID}, clipped to this interval, is the point forecast. It occupies the output field named \texttt{q50}, which contains the requested median for prompted models. The fine-tuned point forecast is generated separately from the bin probabilities. Appendix~\ref{app:metrics} explains how both output formats are scored.

\subsection{Models}
\label{app:models}

\paragraph{Inference.}
All models decode greedily, and the GPT-6 models run with low reasoning effort.

\paragraph{Fine-Tuning.}
We fine-tune all parameters with LLaMA-Factory~\citep{zheng2024llamafactory} for three epochs, with a cosine schedule, 3\% warm-up, and an effective batch size of 32. The learning rate is $2\times10^{-5}$ for Qwen3-1.7B and $1\times10^{-5}$ for Qwen3-8B.
Each fine-tuning sample uses $K=0$ or $K=3$, assigned by a hash of its identifier. The Polymarket fine-tuning data contain 80\% anchored windows and 20\% control windows, with horizons of 1\,h (70\%), 5\,min, and 24\,h (15\% each). Training and inference use matching system prompts and output formats, with thinking disabled.

\paragraph{Training Seeds.}
Each fine-tuned result comes from one training run. To gauge seed variation, we train Qwen3-1.7B on 20k Diplomacy transitions with three seeds (Table~\ref{tab:seeds}). Every score has a standard deviation below one percentage point, and SF outperforms Task~S in all three runs.

\begin{table}[h]
\centering
\footnotesize
\setlength{\tabcolsep}{3pt}
\resizebox{\columnwidth}{!}{%
\begin{tabular}{@{}lccccc@{}}
\toprule
Seed & J & S & SF & SF+engine & Top-1 \\
\midrule
42 & 59.7 & 20.1 & 22.2 & 23.1 & 47.6 \\
43 & 59.6 & 19.6 & 21.5 & 22.1 & 46.8 \\
44 & 60.7 & 19.9 & 23.1 & 23.4 & 48.1 \\
\midrule
Mean (sd) & 60.0 (0.6) & 19.8 (0.3) & 22.2 (0.8) & 22.9 (0.7) & 47.5 (0.7) \\
\bottomrule
\end{tabular}}
\caption{Qwen3-1.7B fine-tuned on 20k Diplomacy transitions with three training seeds (\%, EM except Top-1). SF first predicts the other players' orders, then the next board, and SF+engine resolves its predicted orders in the engine (Appendix~\ref{app:actors}). Top-1 is the share of the other players' orders that the SF model predicts correctly, computed per sample and averaged over samples, with samples in which the other players issue no orders counted as correct. Per-order accuracy (Table~\ref{tab:opp}) instead pools all units.}
\label{tab:seeds}
\end{table}

\paragraph{Compute.}
Fine-tuning and inference for models up to 4B parameters ran on NVIDIA RTX PRO 6000 GPUs with 96\,GB of memory, and Qwen3-8B was fine-tuned on four NVIDIA GH200 or H200 GPUs. The experiments used about 550 GPU hours: about 230 for fine-tuning on the RTX PRO 6000 GPUs, about 230 for fine-tuning Qwen3-8B, and about 90 for inference. The two closed models were queried through their API.

\subsection{Baselines}
\label{app:baselines}

\paragraph{Diplomacy.}
\emph{No change} copies the current board. \emph{Orders applied literally} moves every unit to its ordered destination, retreats dislodged units as ordered, applies builds and disbands, and transfers supply centers by occupancy at the end of a year. It does not adjudicate conflicts. When two units claim one province, the first in player order takes it and the other stays. In Task~S, this baseline leaves the other players' units in place. Three further Task~S baselines take the other players' orders from a uniformly random legal profile, the most frequent order of each unit in the training split, or DipNet~\citep{paquette2019no}. The engine then resolves the joint action.

\paragraph{Polymarket.}
\emph{No change} predicts a zero change, with the training-split rate of no change as its probability. We train LightGBM~\citep{ke2017lightgbm} on the full training split for each task and horizon. Its inputs are 25 hand-crafted features (prices, volume, recent flow, market age and time to close, the anchor order, the agent's position, topic) and aggregates of the window's orders. These aggregates cover all orders in Task~J and the agent's own in Task~S. A classifier predicts the no-change probability, and three quantile regressors predict the change (300 trees, learning rate 0.05, 63 leaves).

\subsection{Metrics}
\label{app:metrics}

\paragraph{Diplomacy.}
EM requires every player's units and supply centers to match the record. Movement phases count as high-interaction because all players' moves, supports, and convoys are adjudicated jointly. Retreat and adjustment phases count as low-interaction because each order there has a direct and almost always independent effect.

\paragraph{Polymarket.}
For Acc, we classify a window as unchanged when \texttt{p\_zero} is at least 0.5. Otherwise, we use the sign of the point forecast, falling back to the change implied by the predicted buy and sell prices when that forecast is zero. Acc is the share of all windows classified correctly. Direction accuracy (Dir) uses the same point-forecast sign and fallback on windows where the price actually changes, without applying the \texttt{p\_zero} threshold.

For CRPS, we construct a distribution from each model's outputs in the same way. Let $l$, $m$, and $u$ be the sorted lower percentile, point forecast, and upper percentile. We interpolate a piecewise linear CDF through $(l,0.1)$, $(m,0.5)$, and $(u,0.9)$. It reaches zero at $l-2\max(m-l,10^{-4})$ and one at $u+2\max(u-m,10^{-4})$. We integrate its squared distance from the observed outcome's step CDF on $[-1,1]$ and report CRPS in cents. In the single-step tables, unparseable predictions count as incorrect for accuracy and are excluded from CRPS.

\section{Prediction Given All Actors' Actions (Q1)}
\label{app:q1}

\subsection{Full Single-Step Results}
\label{app:l1-full}

Tables~\ref{tab:l1-diplomacy} and~\ref{tab:l1-polymarket} extend Table~\ref{tab:l1} with few-shot prompting and with three previous transitions (Diplomacy) or hours (Polymarket) as history ($K=3$). Among prompted models in Diplomacy Task~J, only zero-shot GPT-6-Sol at $K=3$ has a higher EM than applying the orders literally (Table~\ref{tab:l1-diplomacy}). In Polymarket, every prompted model has a higher CRPS than the no-change forecast in every setting (Table~\ref{tab:l1-polymarket}). Appendix~\ref{app:leak} tests whether prompted models know the market outcomes.

\begin{table*}[h]
\centering
\small
\setlength{\tabcolsep}{4pt}
\begin{tabular}{@{}llcccccccc@{}}
\toprule
& & \multicolumn{4}{c}{$K=0$} & \multicolumn{4}{c}{$K=3$} \\
\cmidrule(lr){3-6}\cmidrule(l){7-10}
Setting & Model & EM & EM$_\mathrm{high}$ & EM$_\mathrm{low}$ & F1 & EM & EM$_\mathrm{high}$ & EM$_\mathrm{low}$ & F1 \\
\midrule
\multicolumn{10}{c}{\textit{Task J (joint action)}} \\
\midrule
Zero-shot & GPT-6-Luna & 29.1 & 3.4 & 55.2 & 90.8 & 27.8 & 3.6 & 52.4 & 89.8 \\
 & GPT-6-Sol & 40.5 & 10.1 & 71.4 & 96.3 & 53.3 & 18.8 & 88.3 & 96.5 \\
\cmidrule(l){2-10}
Few-shot (3-shot) & GPT-6-Luna & 27.6 & 1.6 & 54.0 & 89.8 & 23.2 & 1.4 & 45.4 & 89.1 \\
 & GPT-6-Sol & 43.8 & 12.1 & 76.0 & 96.0 & 49.7 & 14.7 & 85.3 & 96.6 \\
\cmidrule(l){2-10}
SFT (20k) & Qwen3-1.7B & 59.7 & 23.6 & 96.4 & 96.9 & 60.7 & 23.8 & 98.2 & 97.0 \\
 & Qwen3-8B & \textbf{65.1} & \textbf{30.8} & \textbf{100.0} & \textbf{97.7} & \textbf{64.5} & \textbf{30.4} & \textbf{99.2} & \textbf{97.8} \\
\cmidrule(l){2-10}
Baseline & No change & 1.7 & 1.0 & 2.4 & 81.5 & 1.7 & 1.0 & 2.4 & 81.5 \\
 & Orders applied literally & 51.1 & 4.8 & 98.2 & 92.1 & 51.1 & 4.8 & 98.2 & 92.1 \\
\midrule
\multicolumn{10}{c}{\textit{Task S (own action only)}} \\
\midrule
Zero-shot & GPT-6-Luna & 6.7 & 1.0 & 12.5 & 82.6 & 5.1 & 0.9 & 9.3 & 82.7 \\
 & GPT-6-Sol & 7.5 & 1.4 & 13.7 & 83.3 & 11.2 & 1.6 & 21.1 & 83.3 \\
\cmidrule(l){2-10}
Few-shot (3-shot) & GPT-6-Luna & 5.7 & 1.0 & 10.4 & 82.5 & 4.5 & 0.8 & 8.2 & 82.5 \\
 & GPT-6-Sol & 14.1 & 1.7 & 26.8 & 83.5 & 12.7 & 1.3 & 24.2 & 83.8 \\
\cmidrule(l){2-10}
SFT (20k) & Qwen3-1.7B & 20.1 & 1.6 & 38.8 & 84.1 & 20.8 & 1.6 & 40.3 & 84.3 \\
 & Qwen3-8B & \textbf{20.8} & \textbf{1.8} & \textbf{40.0} & \textbf{84.3} & \textbf{21.1} & \textbf{1.8} & \textbf{40.6} & \textbf{84.4} \\
\cmidrule(l){2-10}
Baseline & Own orders applied literally & 7.7 & 1.9 & 13.6 & 83.0 & 7.7 & 1.9 & 13.6 & 83.0 \\
 & Uniform legal + engine & 8.5 & 0.2 & 16.9 & -- & 8.5 & 0.2 & 16.9 & -- \\
 & Most frequent + engine & 8.2 & 0.1 & 16.5 & -- & 8.2 & 0.1 & 16.5 & -- \\
 & DipNet + engine & 24.7 & 0.1 & 49.6 & -- & 24.7 & 0.1 & 49.6 & -- \\
\bottomrule
\end{tabular}
\caption{Single-step prediction in seven-player Diplomacy (\%) on the shared subset (1,000 transitions, 2,000 Task~S samples). $K$ is the number of previous transitions given as history. EM$_\mathrm{high}$ and EM$_\mathrm{low}$ are restricted to high- and low-interaction transitions. Fine-tuned models are evaluated zero-shot, and baselines use no history. In the ``+ engine'' rows, the named policy supplies the other players' orders and the engine resolves the joint action (Appendix~\ref{app:baselines}). Bold marks the best prompted or fine-tuned LLM per column, baselines excluded.}
\label{tab:l1-diplomacy}
\end{table*}

\begin{table*}[h]
\centering
\small
\setlength{\tabcolsep}{3.5pt}
\begin{tabular}{@{}llcccccccccc@{}}
\toprule
& & \multicolumn{5}{c}{$K=0$} & \multicolumn{5}{c}{$K=3$} \\
\cmidrule(lr){3-7}\cmidrule(l){8-12}
Setting & Model & Acc & Acc$_\mathrm{high}$ & Acc$_\mathrm{low}$ & Dir & CRPS & Acc & Acc$_\mathrm{high}$ & Acc$_\mathrm{low}$ & Dir & CRPS \\
\midrule
\multicolumn{12}{c}{\textit{Task J (joint action)}} \\
\midrule
Zero-shot & GPT-6-Luna & 43.0 & 36.8 & 48.0 & 56.2 & 0.42 & 45.9 & 40.6 & 50.2 & 57.2 & 0.45 \\
 & GPT-6-Sol & 54.6 & 48.7 & 59.5 & 56.5 & 0.55 & 55.4 & 51.3 & 58.8 & 57.6 & 0.63 \\
\cmidrule(l){2-12}
Few-shot (3-shot) & GPT-6-Luna & 44.1 & 41.1 & 46.5 & 56.4 & 0.41 & 41.6 & 37.1 & 45.3 & 56.6 & 0.39 \\
 & GPT-6-Sol & 52.7 & 52.3 & 53.1 & 58.4 & 0.49 & 52.5 & 51.3 & 53.5 & 59.0 & 0.43 \\
\cmidrule(l){2-12}
SFT (20k) & Qwen3-1.7B & \textbf{66.1} & \textbf{62.5} & 69.0 & \textbf{62.3} & \textbf{0.29} & 67.0 & 63.2 & 70.1 & 63.1 & 0.29 \\
 & Qwen3-8B & 65.5 & 60.5 & \textbf{69.5} & 61.4 & \textbf{0.29} & \textbf{67.7} & \textbf{63.5} & \textbf{71.2} & \textbf{64.3} & \textbf{0.27} \\
\cmidrule(l){2-12}
Baseline & No change & 17.8 & 2.0 & 30.6 & -- & 0.37 & 17.8 & 2.0 & 30.6 & -- & 0.37 \\
 & LightGBM & 68.8 & 67.1 & 70.2 & 65.9 & 0.30 & 68.8 & 67.1 & 70.2 & 65.9 & 0.30 \\
\midrule
\multicolumn{12}{c}{\textit{Task S (own action only)}} \\
\midrule
Zero-shot & GPT-6-Luna & 46.1 & 40.3 & 50.9 & 56.7 & 0.48 & 43.6 & 37.6 & 48.4 & 55.5 & 0.49 \\
 & GPT-6-Sol & 48.5 & 42.3 & 53.5 & 55.8 & 0.47 & 48.5 & 43.0 & 53.1 & 56.5 & 0.52 \\
\cmidrule(l){2-12}
Few-shot (3-shot) & GPT-6-Luna & 44.8 & 43.1 & 46.1 & 57.0 & 0.54 & 44.3 & 39.5 & 48.2 & 56.9 & 0.58 \\
 & GPT-6-Sol & 49.9 & 46.5 & 52.7 & 57.8 & 0.85 & 45.2 & 40.5 & 49.1 & 56.6 & 0.65 \\
\cmidrule(l){2-12}
SFT (20k) & Qwen3-1.7B & 59.9 & 55.2 & \textbf{63.7} & 59.7 & \textbf{0.31} & 61.1 & 56.2 & \textbf{65.0} & 59.9 & \textbf{0.31} \\
 & Qwen3-8B & \textbf{60.2} & \textbf{55.9} & \textbf{63.7} & \textbf{59.9} & 0.32 & \textbf{62.4} & \textbf{60.4} & 64.1 & \textbf{61.9} & \textbf{0.31} \\
\cmidrule(l){2-12}
Baseline & No change & 17.8 & 2.0 & 30.6 & -- & 0.37 & 17.8 & 2.0 & 30.6 & -- & 0.37 \\
 & LightGBM & 61.4 & 58.2 & 63.9 & 61.7 & 0.31 & 61.4 & 58.2 & 63.9 & 61.7 & 0.31 \\
\bottomrule
\end{tabular}
\caption{Single-step prediction in Polymarket ($\Delta = 1$\,h, accuracies in \%) on the shared subset (1,333 decision points). $K$ is the number of previous hours given as history. Acc$_\mathrm{high}$ and Acc$_\mathrm{low}$ are restricted to high- and low-interaction windows. Dir is the direction accuracy on windows in which the mid-price changes. CRPS is in cents (lower is better). Fine-tuned models are evaluated zero-shot, and baselines use no history. Bold marks the best prompted or fine-tuned LLM per column, baselines excluded.}
\label{tab:l1-polymarket}
\end{table*}

\subsection{Scaling Results}
\label{app:scale}

Table~\ref{tab:scaling} gives the numbers behind Figure~\ref{fig:scale}. The 5k, 80k, and 320k training sets have the composition of the 20k set (Appendix~\ref{app:models}). They are nested in Diplomacy and drawn independently from the training split in Polymarket. All models follow the recipe of the 20k models, with the same number of epochs, so larger sets take more optimizer steps. The 0.6B and 4B models use the learning rate of Qwen3-1.7B.

As shown in Table~\ref{tab:scaling}, in Polymarket, where the observation already carries recent trades, more data raises Task~S nearly as much as Task~J. On high-interaction windows, however, the gap between the two tasks widens with data from 3 to 10 points. In Diplomacy, the modest Task~S gains with data come from low-interaction transitions. Hyperparameters are not tuned per size, which may explain why the 8B model falls below the 4B model in Task~J in both worlds.

\begin{table}[h]
\centering
\footnotesize
\setlength{\tabcolsep}{2pt}
\resizebox{\columnwidth}{!}{%
\begin{tabular}{@{}llcccc@{}}
\toprule
& & \multicolumn{2}{c}{Diplomacy (EM)} & \multicolumn{2}{c}{Polymarket (Acc)} \\
\cmidrule(lr){3-4}\cmidrule(l){5-6}
Axis & Setting & J & S & J & S \\
\midrule
Data (1.7B) & 5k & 48.9 (11.3) & 16.2 (0.8) & 62.0 (56.5) & 56.6 (53.2) \\
 & 20k & 59.7 (23.6) & 20.1 (1.6) & 66.1 (62.5) & 59.9 (55.2) \\
 & 80k & 74.7 (49.8) & 23.0 (1.9) & 69.5 (66.7) & 61.8 (55.4) \\
 & 320k & 88.4 (77.0) & 25.4 (1.9) & 71.6 (69.1) & 64.9 (58.9) \\
\midrule
Size (20k) & 0.6B & 57.6 (19.2) & 20.5 (1.6) & 63.3 (59.2) & 58.1 (53.8) \\
 & 1.7B & 59.7 (23.6) & 20.1 (1.6) & 66.1 (62.5) & 59.9 (55.2) \\
 & 4B & 66.2 (33.7) & 20.5 (1.8) & 67.2 (63.7) & 61.5 (56.2) \\
 & 8B & 65.1 (30.8) & 20.8 (1.8) & 65.5 (60.5) & 60.2 (55.9) \\
\midrule
Both & 8B, 320k & 93.8 (87.7) & 24.4 (1.9) & 72.5 (70.4) & 65.2 (61.0) \\
\bottomrule
\end{tabular}}
\caption{Scaling the fine-tuned Qwen3 models in training data and model size (\%, shared subsets). Values on high-interaction transitions are in parentheses.}
\label{tab:scaling}
\end{table}

\subsection{From-Scratch Control}
\label{app:scratch}

We train Qwen3-1.7B from random initialization on the same 20k and 320k transitions as the fine-tuned model. Training follows the recipe of Appendix~\ref{app:models}, except for a learning rate of $10^{-4}$, chosen on the Diplomacy validation games.

As shown in Table~\ref{tab:scratch}, at 320k transitions the model trained from scratch matches the pretrained model on low-interaction Diplomacy transitions in Task~J, so on these transitions the log alone suffices. On high-interaction transitions, however, it learns far less than the pretrained model from the same data, and the gap widens with data. It also fails on the two-player games because it keeps writing seven-player boards. In Polymarket, the model trained from scratch on 320k transitions reaches about the Task~J Acc of the pretrained model trained on 20k. However, with the same data, the pretrained model stays ahead, and by more on the later markets.

In Task~S, the two initializations differ less (Table~\ref{tab:scratch}). In Diplomacy they are close at 320k. At 20k the pretrained model is ten points ahead, while the model trained from scratch also remains much less accurate on low-interaction Task~J transitions. In Polymarket, the pretrained model's Task~S margin is smaller than its Task~J margin. Pretraining therefore brings larger gains when the joint action is known. Prediction from only the agent's action remains difficult under both initializations.

Pretraining also helps on actions never taken, but much less at 320k. On the alternative orders in Diplomacy (Appendix~\ref{app:l2l3}), evaluated in Task~S form on the common evaluation set, the pretrained model's F1 on board changes is 43.9\% against 22.1\% from scratch at 20k, and 47.1\% against 44.1\% at 320k. More data thus raises the pretrained model's F1 far less than its Task~J EM.

\begin{table}[h]
\centering
\footnotesize
\setlength{\tabcolsep}{3pt}
\resizebox{\columnwidth}{!}{%
\begin{tabular}{@{}llccccccccc@{}}
\toprule
& & \multicolumn{5}{c}{Diplomacy (EM)} & \multicolumn{3}{c}{Polymarket (Acc)} \\
\cmidrule(lr){3-7}\cmidrule(l){8-10}
Data & Initialization & J & J$_\mathrm{high}$ & J$_\mathrm{low}$ & J, 2p & S & J & J, later & S \\
\midrule
20k & pretrained & 59.7 & 23.6 & 96.4 & 61.1 & 20.1 & 66.1 & 67.4 & 59.9 \\
 & random & 32.0 & 4.6 & 59.9 & 0.3 & 10.2 & 60.8 & 59.6 & 57.8 \\
\midrule
320k & pretrained & 88.4 & 77.0 & 100.0 & 88.4 & 25.4 & 71.6 & 74.0 & 64.9 \\
 & random & 66.2 & 32.9 & 100.0 & 11.0 & 24.1 & 67.5 & 65.4 & 61.7 \\
\bottomrule
\end{tabular}}
\caption{The Qwen3-1.7B architecture, fine-tuned from the pretrained checkpoint or trained from random initialization (\%, $K=0$). J$_\mathrm{high}$, J$_\mathrm{low}$: high- and low-interaction transitions. J, 2p: Task~J on the two-player common evaluation set. J, later: Task~J on 1,333 decision points of the later markets. The other columns use the shared subsets.}
\label{tab:scratch}
\end{table}

\subsection{Prediction at Different Horizons}
\label{app:horizons}
The horizon probe of Figure~\ref{fig:l2-clock}a evaluates the 1,333 decision points of the Polymarket shared subset at 5 minutes, 1 hour, and 24 hours, with the orders of each corresponding window. All three horizons appear in fine-tuning. We compare the mean predicted probability of no price change with its observed frequency at each horizon.

\subsection{Robustness Results}
\label{app:robust}

Table~\ref{tab:robust} evaluates the fine-tuned models of Table~\ref{tab:l1} on the common evaluation sets. Its Task~J rows support the robustness claims of Section~\ref{sec:q1}. In Task~S, the margin over applying the agent's own orders literally is similar on two-player and seven-player Diplomacy games. On high-interaction transitions, however, Task~S is no better than this baseline in either. In Polymarket, Task~S stays well below Task~J on high-interaction windows in both periods.

\begin{table}[h]
\centering
\small
\setlength{\tabcolsep}{4pt}
\begin{tabular}{@{}lcccccc@{}}
\toprule
 & \multicolumn{2}{c}{Qwen3-1.7B} & \multicolumn{2}{c}{Qwen3-8B} & \multicolumn{2}{c}{Baseline} \\
\cmidrule(lr){2-3}\cmidrule(lr){4-5}\cmidrule(lr){6-7}
Split & All & High & All & High & All & High \\
\midrule
\multicolumn{7}{@{}l}{\emph{Diplomacy, Task~J}} \\
7p & 60.9 & 22.9 & 65.7 & 30.7 & 52.0 & 4.0 \\
2p & 61.1 & 38.6 & 69.6 & 53.0 & 53.1 & 23.9 \\
\midrule
\multicolumn{7}{@{}l}{\emph{Diplomacy, Task~S}} \\
7p & 19.7 & 1.2 & 19.8 & 1.2 & 7.3 & 1.5 \\
2p & 22.2 & 0.7 & 22.4 & 0.9 & 9.4 & 1.0 \\
\midrule
\multicolumn{7}{@{}l}{\emph{Polymarket, Task~J}} \\
test & 66.8 & 62.1 & 66.8 & 62.6 & 68.1 & 63.9 \\
later & 67.5 & 64.8 & 68.8 & 65.9 & 65.7 & 63.6 \\
\midrule
\multicolumn{7}{@{}l}{\emph{Polymarket, Task~S}} \\
test & 61.4 & 57.4 & 62.2 & 57.9 & 62.7 & 58.5 \\
later & 61.1 & 57.9 & 62.1 & 59.0 & 60.9 & 60.4 \\
\bottomrule
\end{tabular}
\caption{Robustness on the common evaluation sets of Appendix~\ref{app:data} (EM in Diplomacy, Acc in Polymarket, \%). High: high-interaction transitions. 7p: seven-player games. 2p: two-player games, unseen in training. Test: the in-distribution test split. Later: the later-market split. Baselines: orders applied literally in Diplomacy (in Task~S only the agent's own), LightGBM in Polymarket.}
\label{tab:robust}
\end{table}

\subsection{Outcome Knowledge of the Prompted Models}
\label{app:leak}

The GPT-6 models may know how Polymarket questions resolved. We show each model only a market's question, event, and end date. In Table~\ref{tab:leak}, both models say they know the outcome of most test-period markets and are correct in 97.9\% (GPT-6-Luna) and 98.8\% (GPT-6-Sol) of those cases. They make few such claims about later markets. Despite this drop in reported outcome knowledge, their one-hour prediction accuracy does not fall on later markets, and changes by amounts similar to those of the fine-tuned models. This comparison does not test knowledge of when news arrived within an hour.

\begin{table}[h]
\centering
\scriptsize
\setlength{\tabcolsep}{2.5pt}
\begin{tabular}{@{}lcccc@{}}
\toprule
& \multicolumn{2}{c}{Outcome known} & \multicolumn{2}{c}{Acc, test $\to$ later} \\
\cmidrule(lr){2-3}\cmidrule(l){4-5}
Model & Test & Later & Task~J & Task~S \\
\midrule
GPT-6-Luna & 81.9 & 1.6 & 43.0 $\to$ 46.7 & 46.1 $\to$ 48.5 \\
GPT-6-Sol & 68.1 & 0.0 & 54.6 $\to$ 57.5 & 48.5 $\to$ 52.6 \\
Qwen3-1.7B (SFT) & -- & -- & 66.1 $\to$ 67.4 & 59.9 $\to$ 61.4 \\
Qwen3-8B (SFT) & -- & -- & 65.5 $\to$ 70.7 & 60.2 $\to$ 63.1 \\
LightGBM & -- & -- & 68.8 $\to$ 67.1 & 61.4 $\to$ 61.8 \\
\bottomrule
\end{tabular}
\caption{Left: share (\%) of resolved markets whose outcome each GPT-6 model says it knows (238 test and 440 later markets). Among test markets whose outcomes the models say they know, accuracy is 97.9\% for GPT-6-Luna and 98.8\% for GPT-6-Sol. Right: Task~J and Task~S Acc (\%) on the 1,333-point subsets of the in-distribution test split and the later-market split.}
\label{tab:leak}
\end{table}

\section{Prediction Given Only the Agent's Actions (Q2)}
\label{app:q2}

\subsection{History-Length Sweep}
\label{app:l2-history}

Table~\ref{tab:l2-history} gives the full history sweep behind Figure~\ref{fig:hist}. The input adds the previous $K$ transitions (Diplomacy) or hours of trades (Polymarket), for $K \in \{0, 1, 3, 6, 10\}$. In Polymarket, $K$ hours of history consist of $K$ hourly price and flow bars and the most recent orders in those hours, at most 20 for $K \le 3$ and $7K$ beyond. We evaluate GPT-6-Sol and GPT-6-Luna zero-shot and a Qwen3-1.7B fine-tuned on the 20k samples of Table~\ref{tab:l1}, with $K$ drawn uniformly from the same values. This checkpoint is trained separately from the Qwen3-1.7B of Table~\ref{tab:l1} (Appendix~\ref{app:models}), so its scores at $K=0$ differ slightly.

For the fine-tuned model and GPT-6-Sol, history helps Diplomacy Task~S only on low-interaction transitions. On high-interaction transitions, their Task~S EM stays below 2\% at every $K$ (Table~\ref{tab:l2-history}). In Polymarket, where the observation already summarizes recent trades, history changes Task~S only slightly. History also does not close the gap between the two tasks. The gap stays flat for the fine-tuned model in both worlds. For GPT-6-Sol it widens in Diplomacy, because its Task~J gains more from history than its Task~S.

The fine-tuned model's Task~J changes little with history in either world. GPT-6-Sol benefits from history in Diplomacy Task~J, while GPT-6-Luna becomes less accurate. In Polymarket, history changes the prompted models' Task~J by a few points at most.

\begin{table*}[h]
\centering
\small
\setlength{\tabcolsep}{4pt}
\begin{tabular}{@{}llccccccccccc@{}}
\toprule
& & \multicolumn{5}{c}{Diplomacy (EM)} & & \multicolumn{5}{c}{Polymarket (Acc)} \\
\cmidrule(lr){3-7}\cmidrule(l){9-13}
Model & Task & $K=0$ & $K=1$ & $K=3$ & $K=6$ & $K=10$ & & $K=0$ & $K=1$ & $K=3$ & $K=6$ & $K=10$ \\
\midrule
Qwen3-1.7B (SFT) & J & 58.9 & 59.4 & 59.6 & 60.1 & 59.7 &  & 65.5 & 66.5 & 66.8 & 66.2 & 66.5 \\
 & S & 19.9 & 22.2 & 21.9 & 21.1 & 20.8 &  & 60.5 & 60.8 & 61.9 & 61.8 & 61.1 \\
 & S, low & 38.8 & 43.6 & 42.8 & 41.2 & 40.7 &  & 63.8 & 63.8 & 65.0 & 64.9 & 63.7 \\
 & S, high & 1.3 & 1.2 & 1.3 & 1.2 & 1.2 &  & 56.5 & 57.2 & 58.0 & 58.0 & 58.0 \\
\midrule
GPT-6-Sol & J & 40.5 & 47.4 & 53.3 & 54.5 & 51.7 &  & 54.6 & 56.3 & 55.4 & 54.1 & 54.8 \\
 & S & 7.5 & 11.9 & 11.2 & 10.8 & 9.8 &  & 48.5 & 48.1 & 48.5 & 48.1 & 48.5 \\
 & S, low & 13.7 & 22.6 & 21.1 & 20.1 & 18.2 &  & 53.5 & 54.0 & 53.1 & 53.1 & 53.3 \\
 & S, high & 1.4 & 1.4 & 1.6 & 1.7 & 1.6 &  & 42.3 & 40.8 & 43.0 & 42.0 & 42.6 \\
\midrule
GPT-6-Luna & J & 29.1 & 29.5 & 27.8 & 24.6 & 22.7 &  & 43.0 & 44.3 & 45.9 & 45.9 & 47.0 \\
 & S & 6.7 & 5.5 & 5.1 & 3.9 & 3.4 &  & 46.1 & 45.3 & 43.6 & 44.6 & 44.4 \\
 & S, low & 12.5 & 10.1 & 9.3 & 7.0 & 6.2 &  & 50.9 & 50.7 & 48.4 & 48.6 & 47.9 \\
 & S, high & 1.0 & 1.0 & 0.9 & 0.8 & 0.5 &  & 40.3 & 38.6 & 37.6 & 39.8 & 40.1 \\
\bottomrule
\end{tabular}
\caption{History-length sweep on the shared subsets (EM in Diplomacy and Acc in Polymarket, in \%). All models are evaluated zero-shot. $K$ is the number of previous transitions or hours in the input. ``S, low'' and ``S, high'' restrict Task~S to low- and high-interaction transitions.}
\label{tab:l2-history}
\end{table*}

\subsection{Predicting the Other Actors' Behavior}
\label{app:actors}
The factored variant (SF) receives the Task~S input and adds predictions of the other actors' behavior. We query GPT-6 zero-shot with modified output instructions. In Diplomacy, the prompt asks for the other players' orders in an \texttt{OTHERS} block, followed by \texttt{NEXT} and \texttt{STATE}. In Polymarket, the requested JSON schema lists the price fields followed by an \texttt{others} object. Its keys \texttt{N}, \texttt{F}, and \texttt{V} represent the other traders' order count, net flow, and volume over the hour.

We fine-tune Qwen3-1.7B and, in Diplomacy only, Qwen3-8B as SF models. They use the same recipe and 20k inputs as the fine-tuned models of each world. Their Task~S targets place the other actors' behavior before the next observation.

For Diplomacy we score both the model's own next board (SF) and the engine's resolution of the agent's recorded orders and the predicted orders (SF+engine). The latter removes errors in resolving a joint action. In Table~\ref{tab:l2-gap-full}, replacing the fine-tuned models' board predictions with engine outcomes helps little. It helps the prompted models more, indicating that their errors also involve resolving the predicted orders.

\begin{table*}[h]
\centering
\small
\setlength{\tabcolsep}{4pt}
\begin{tabular}{@{}lccccccccc@{}}
\toprule
& \multicolumn{5}{c}{Diplomacy (EM)} & \multicolumn{4}{c}{Polymarket (Acc)} \\
\cmidrule(lr){2-6}\cmidrule(l){7-10}
Model & J & S & SF & SF+engine & Share & J & S & SF & Share \\
\midrule
GPT-6-Luna (zero-shot) & 29.1 & 6.7 & 9.3 & 17.2 & 0.24 & 43.0 & 46.1 & 47.6 & $-$0.06 \\
GPT-6-Sol (zero-shot) & 40.5 & 7.5 & 11.9 & 17.1 & 0.36 & 54.6 & 48.5 & 52.6 & 0.12 \\
Qwen3-1.7B (SFT) & 59.7 & 20.1 & 22.2 & 23.1 & 0.50 & 66.1 & 59.9 & 60.4 & 0.16 \\
Qwen3-8B (SFT) & 65.1 & 20.8 & 22.9 & 23.4 & 0.56 & 65.5 & 60.2 & -- & 0.13 \\
\midrule
DipNet policy + engine & -- & -- & -- & 24.7 & -- & -- & -- & -- & -- \\
\bottomrule
\end{tabular}
\caption{Prediction with and without the other actors' actions. SF adds predictions of their behavior to the Task~S output, and SF+engine resolves its predicted orders in the engine. Share $=(\mathrm{Err}_S-\mathrm{Err}_J)/\mathrm{Err}_S$, with Err the error rate, is the relative reduction in error rate from Task~S to Task~J.}
\label{tab:l2-gap-full}
\end{table*}

\paragraph{Per-Order Accuracy on the Other Players' Orders.}
Whole-board EM can remain low even as individual order predictions improve. We therefore also report per-order accuracy, the share of the other players' units whose predicted order matches the recorded one (Table~\ref{tab:opp}). A unit is interacting if the provinces its order acts on overlap those of another player's unit, with a unit without an order treated as holding. On interacting units, the fine-tuned Qwen3-1.7B beats GPT-6-Sol, which does no better than the most frequent order of each unit. Given three phases of history ($K=3$), the three seed checkpoints of Table~\ref{tab:seeds} predict the other players' orders more accurately, but their high-interaction EM stays below 2\%. DipNet, trained on far more human games, has the highest per-order accuracy, though its whole-board EM on high-interaction transitions is only 0.1\%.

\begin{table}[h]
\centering
\footnotesize
\setlength{\tabcolsep}{4pt}
\resizebox{\columnwidth}{!}{%
\begin{tabular}{@{}lccc@{}}
\toprule
Model & All units & Interacting & EM$_\mathrm{high}$ \\
\midrule
Most frequent order & 22.6 & 16.2 & 0.1 \\
GPT-6-Luna (zero-shot) & 16.8 & 10.9 & 1.5 \\
GPT-6-Sol (zero-shot) & 20.7 & 16.5 & 0.1 \\
Qwen3-1.7B (SFT, 3 seeds) & 29.6 & 25.5 & 1.9 \\
Qwen3-8B (SFT) & 30.7 & 26.6 & 1.8 \\
Qwen3-1.7B (SFT), $K=3$ & 36.1 & 31.9 & 1.8 \\
DipNet & 45.9 & 43.0 & 0.1 \\
\bottomrule
\end{tabular}}
\caption{Per-order accuracy (\%) of the SF models and two baselines on the other players' orders in the 2,000 Task~S samples of the shared subset, over all units and over interacting units. EM$_\mathrm{high}$ is the high-interaction EM after the engine resolves the predicted orders.}
\label{tab:opp}
\end{table}

\subsection{Negotiation Messages}
\label{app:negotiation}
Finding~2 motivates testing whether messages help predict other players' actions. We repeat Task~S in press games, where players exchange messages, and give the model the messages the agent sent or received in the phase. We use 24,925 engine-verified games and test on 1,000 transitions with two players each as the agent. Of these samples, 44\% have no message. We fine-tune four Qwen3-1.7B models on 20k press transitions, using Task~S or SF with and without messages.

In Table~\ref{tab:press}, messages do not improve the fine-tuned Task~S model's EM and raise the SF model's Top-1 by about one point. GPT-6-Sol's EM rises with messages, but also rises when their text is blanked. The remaining gain is on low-interaction transitions. Adding raw negotiation messages therefore brings little improvement in this setup.

\begin{table}[h]
\centering
\scriptsize
\setlength{\tabcolsep}{2.5pt}
\begin{tabular}{@{}lccccc@{}}
\toprule
& \multicolumn{3}{c}{Task~S EM} & \multicolumn{2}{c}{Others' orders} \\
\cmidrule(lr){2-4}\cmidrule(l){5-6}
Model & none & messages & blank & SF & SF + msg. \\
\midrule
Qwen3-1.7B, press-tuned & 18.3 & 18.2 & 17.6 & 47.0 & 47.9 \\
GPT-6-Sol & 8.9 & 11.1 & 10.2 & 31.8 & 32.9 \\
GPT-6-Luna & 6.8 & 6.9 & 7.1 & 26.3 & 26.2 \\
\bottomrule
\end{tabular}
\caption{Task~S in press games (\%, prompted models zero-shot). EM is given without the agent's messages, with them, and with their text blanked. Others' orders gives SF Top-1 (Table~\ref{tab:seeds}), without and with messages.}
\label{tab:press}
\end{table}

\subsection{Idle Agents}
\label{app:idle}
In the idle probe of Figure~\ref{fig:l2-clock}b (Section~\ref{sec:q2}), the idle cases in Polymarket are the 333 control windows of the shared subset. In Diplomacy they are the 619 of the 2,000 Task~S samples whose agent has no orders.

In Polymarket, knowing the other traders' orders raises the fine-tuned Qwen3-1.7B's Acc by about 22 points when the agent is idle (Figure~\ref{fig:l2-clock}b). When the agent places a large order, the gain is only about one point. GPT-6-Sol's Acc rises by about six points in both cases. In Diplomacy the idle samples are almost all in retreat and adjustment phases, so only the comparison within a phase type is meaningful (Table~\ref{tab:idle}). In adjustment phases the gain is large in both cases, because the other players issue about four orders per sample, against about one of the agent's when it acts.

\begin{table}[h]
\centering
\footnotesize
\setlength{\tabcolsep}{3pt}
\resizebox{\columnwidth}{!}{%
\begin{tabular}{@{}lcccccc@{}}
\toprule
& \multicolumn{2}{c}{Retreat} & \multicolumn{2}{c}{Adjustment} & \multicolumn{2}{c}{Both} \\
\cmidrule(lr){2-3}\cmidrule(lr){4-5}\cmidrule(l){6-7}
Model & idle & acting & idle & acting & idle & acting \\
\midrule
GPT-6-Luna (zero-shot) & 23.0 & 9.0 & 77.8 & 70.0 & 38.6 & 49.1 \\
GPT-6-Sol (zero-shot) & 35.8 & 17.9 & 92.4 & 91.8 & 51.9 & 66.5 \\
Qwen3-1.7B (SFT) & 45.6 & 18.7 & 80.7 & 82.5 & 55.6 & 60.6 \\
Qwen3-8B (SFT) & 47.7 & 23.9 & 85.4 & 82.5 & 58.4 & 62.4 \\
\bottomrule
\end{tabular}}
\caption{Diplomacy, Task~J minus Task~S EM (points) when the agent issues no orders (idle) or some (acting), by phase type. Retreat phases have 430 idle and 134 acting samples, and adjustment phases 171 and 257.}
\label{tab:idle}
\end{table}

\section{Simulation Beyond the Log (Q3)}
\label{app:q3}

\subsection{Multi-Step Rollouts}
\label{app:rollout}

Table~\ref{tab:l1-rollout-full} gives rollout results for both worlds.

\paragraph{Rollouts in Diplomacy.}
Each of the 500 chains covers five consecutive transitions of a different seven-player test game, starting at a movement phase after the game's first two phases. The model receives its own predicted board from the previous step in the closed loop and the true board in the open loop. In both loops the orders and the phase name of each step come from the record. In Task~S, one player per chain is the agent. A prediction that cannot be parsed ends the chain, and its remaining closed-loop steps score zero.

\paragraph{Rollouts in Polymarket.}
We recursively predict prices while updating trading activity from the recorded log. The 500 chains start at anchored test decision points and run for 24 hourly steps. At each step, the prices come from the previous prediction, while the remaining observation fields come from the log. Task~S gives only the agent's orders for that hour, and Task~J gives all recorded orders. The mid-price moves by the point forecast, and the spread is set from the predicted buy and sell prices. The direct forecast starts from the initial observation and receives the corresponding task's orders for all 24 hours at once.

For rollout Acc, the predicted change is the final rolled-out mid-price minus the starting price. We multiply the 24 hourly no-change scores and classify the rollout as unchanged when the product is at least 0.5. Otherwise, we use the sign of the rolled-out change, with the final buy and sell prices as a fallback. The product is a heuristic score. A path can move and return to its starting price, so the product is not the probability of zero net change over 24 hours. The comparison with the direct forecast tests recursive price prediction with logged activity updates, which the direct forecast does not receive.

GPT-6-Luna's direct forecast has a low Acc because it predicts no change in 45.4\% of the chains in Task~J and 34.2\% in Task~S, while the mid-price does not change over the 24 hours in only 3.6\% of them. On price error, no model beats the no-change forecast, directly or by rollout (mean absolute error of at least 1.02 cents, against 0.99 cents).

\begin{table*}[h]
\centering
\small
\setlength{\tabcolsep}{4pt}
\begin{tabular}{@{}llcccccc@{}}
\toprule
& & \multicolumn{4}{c}{Diplomacy, 5-step chains} & \multicolumn{2}{c}{Polymarket, 24 hours} \\
\cmidrule(lr){3-6}\cmidrule(l){7-8}
Task & Model & EM$_\mathrm{closed}$ & EM$_\mathrm{open}$ & F1$_\mathrm{closed}$ & F1$_\mathrm{open}$ & Acc$_\mathrm{closed}$ & Acc$_\mathrm{direct}$ \\
\midrule
J & Qwen3-1.7B (SFT) & 5.6 & 65.4 & 91.9 & 97.5 & 60.2 & 64.4 \\
J & Qwen3-8B (SFT) & 13.2 & 68.8 & 94.1 & 97.9 & 61.2 & 63.8 \\
J & GPT-6-Luna & 0.0 & 42.2 & 65.7 & 91.7 & 57.4 & 35.6 \\
J & GPT-6-Sol & 0.2 & 57.2 & 83.4 & 96.4 & 57.4 & 55.6 \\
\midrule
S & Qwen3-1.7B (SFT) & 0.0 & 22.0 & 55.1 & 86.0 & 55.6 & 58.6 \\
S & Qwen3-8B (SFT) & 0.0 & 22.4 & 55.4 & 86.5 & 57.0 & 59.0 \\
S & GPT-6-Luna & 0.0 & 9.2 & 52.8 & 85.9 & 56.8 & 43.2 \\
S & GPT-6-Sol & 0.0 & 10.4 & 53.9 & 86.6 & 54.0 & 53.4 \\
\bottomrule
\end{tabular}
\caption{Multi-step prediction (\%), extending Table~\ref{tab:l1-rollout}. Diplomacy scores the board at step 5 with predicted boards (closed loop) or recorded boards (open loop) as inputs. Polymarket recursively predicts prices with logged trading activity (closed) or forecasts 24 hours directly. Always predicting a fall gives 49.4\% Acc.}
\label{tab:l1-rollout-full}
\end{table*}

\subsection{Actions Never Taken in Diplomacy}
\label{app:l2l3}
\label{app:cf}
We use the 504 movement transitions of the shared subset, with two players of each transition as the agent. For each transition and agent, we construct four alternatives to the recorded orders and one control that repeats the recorded orders. The alternatives are a uniformly random legal order set, a set sampled from the training frequencies of each unit's orders, the recorded set with one unit's order changed, and all units holding. This gives \mbox{5,040 order-set evaluations}.

For each alternative, the model predicts the next board for both the recorded and modified orders. We compare the change between its predictions with the change between the two engine outcomes. The engine always holds the other players' recorded orders fixed. Task~S hides these orders, and Task~J includes them in both predictions.

On alternatives that change the next board, F1 compares the predicted and true changes to units and supply centers, and Insensitive is the rate of predicting no change. On alternatives that leave the next board unchanged, Spurious is the rate of predicting a change. Controls are excluded from Spurious because greedy decoding or the GPT-6 cache gives the same prediction for the original orders.

\paragraph{Effect of the Other Players' Orders.}
Giving the other players' orders sharply raises the fine-tuned models' F1 and reduces their spurious changes (Table~\ref{tab:l3-taskj}). The paired F1 gains have standard errors of about one point, clustered by transition. With only the agent's orders, predictions must also account for the unknown orders of other players, which can obscure the effect of changing the agent's action.

The prompted models show no comparable gain. GPT-6-Sol's F1 changes little, while GPT-6-Luna's falls, and both have higher spurious rates in Task~J. The repeated-prompt check below identifies another source of variation.

\paragraph{Repeated-Prompt Check.}
We also check 17 groups of repeated Task~J prompts from the modified-order evaluations. At temperature 0, GPT-6-Sol returns the same change set on every repetition in only 3 groups and GPT-6-Luna in 2. The fine-tuned models do so in all 17. This shows that repeated API calls can produce different predictions even when the joint action is fixed, contributing to spurious changes.

\begin{table*}[t]
\centering
\footnotesize
\setlength{\tabcolsep}{3pt}
\begin{tabular}{@{}lcccccccc@{}}
\toprule
& \multicolumn{4}{c}{Task~S form (other actors imagined)} & \multicolumn{4}{c}{Task~J form (other actors given)} \\
\cmidrule(lr){2-5}\cmidrule(l){6-9}
Model & F1 & Insensitive$\downarrow$ & Spurious$\downarrow$ & Size pred/ref & F1 & Insensitive$\downarrow$ & Spurious$\downarrow$ & Size pred/ref \\
\midrule
GPT-6-Luna (zero-shot) & 47.7 & 6.0 & 45.9 & 3.4/3.9 & 30.1 & 0.1 & 89.7 & 9.5/3.9 \\
GPT-6-Sol (zero-shot) & 56.5 & 4.8 & 40.4 & 3.9/3.9 & 57.9 & 0.3 & 83.3 & 7.6/3.9 \\
Qwen3-1.7B (SFT) & 44.0 & 6.0 & 51.5 & 5.4/3.9 & 73.0 & 3.2 & 33.1 & 4.3/3.9 \\
Qwen3-8B (SFT) & 44.3 & 5.6 & 59.6 & 5.6/3.9 & 77.3 & 2.9 & 29.4 & 4.3/3.9 \\
\bottomrule
\end{tabular}
\caption{Diplomacy action changes, extending Table~\ref{tab:l3}, on 5,040 order-set evaluations from the shared subset (\%, except Size). F1 and Insensitive cover alternatives that change the next board. Spurious covers alternatives that do not, excluding controls that repeat the original orders. Size reports the mean predicted and true numbers of changes to units and centers across all evaluations.}
\label{tab:l3-taskj}
\end{table*}

\subsection{Price-Response Check in Polymarket}
\label{app:pm-response}
On 1,000 anchored test windows whose mid-price lies between 0.2 and 0.8, we remove or reverse all of the agent's orders in the hour. We measure the change from the original to the modified point forecast. Average price responses are larger in this band than near 0 or 1.

The reference comes from a monotone curve fitted on training markets, separately inside and outside this band. It relates average price change in the anchor order's direction to the agent's order volume relative to the previous 24 hours' volume. Reversing the orders reverses the reference response, and removing them sets it to zero. The difference from the response for the original orders gives the reference for each modification. This is an observational comparison with \mbox{average logged responses}.

The first three columns of Table~\ref{tab:l3-full} report order reversal. The GPT-6 models usually predict a response with the same sign as the reference. Fine-tuned models do so in about half of the cases and often predict a small or zero response. The last column uses order removal to test whether larger relative order volumes produce larger predicted effects. Correlations are weak for GPT-6 and Qwen3-1.7B, and near zero for Qwen3-8B.

\begin{table}[h]
\centering
\footnotesize
\setlength{\tabcolsep}{3pt}
\resizebox{\columnwidth}{!}{%
\begin{tabular}{@{}lcccc@{}}
\toprule
Model & Sign & Insensitive$\downarrow$ & Magnitude & Monotonicity \\
\midrule
GPT-6-Luna (zero-shot) & 93.6 & 6.9 & 1.74 & 0.20 \\
GPT-6-Sol (zero-shot) & 84.1 & 15.4 & 2.94 & 0.21 \\
Qwen3-1.7B (SFT) & 52.2 & 41.4 & 0.77 & 0.12 \\
Qwen3-8B (SFT) & 52.3 & 34.2 & 0.93 & $-$0.02 \\
\bottomrule
\end{tabular}}
\caption{Polymarket price-response checks in Task~S (\%, except Magnitude and Monotonicity). The first three columns reverse all of the agent's orders. Sign measures agreement with the nonzero reference response, counting a zero prediction as a miss. Insensitive counts predicted responses smaller than 0.0005 in absolute value. Magnitude is the ratio of mean absolute predicted to reference responses. Monotonicity instead removes the orders and computes Spearman correlation between relative order volume and the predicted effect of including them, with its sign oriented by the anchor order's buy or sell direction.}
\label{tab:l3-full}
\end{table}

\section{Usefulness to Agents (Q4)}
\label{app:agent}

All reported agent evaluations use Diplomacy. Games start from movement phases of the seven-player common evaluation set. Plan consistency uses 500 starting boards, and the other agent experiments share 200. The agent is one player with units, fixed per starting board. Every opponent plays supervised DipNet with greedy decoding. Success means ending the 10 phases with at least the initial number of supply centers.

\subsection{Plan Consistency in Diplomacy}
\label{app:plan}
The agent uses the same greedy supervised DipNet policy as its opponents. In \emph{Real}, each game is played in the engine, and in \emph{WM} it is played inside the world model. At each phase of WM, DipNet orders all players on the world model's board. The world model predicts the next board from the joint orders (Task~J) or from the agent's orders alone (Task~S). A board that cannot be parsed, is physically invalid, or cannot be played by DipNet ends the trajectory as a failure. In \emph{W2R}, the agent's orders from the WM trajectory are replayed in the engine, while the opponents re-plan with DipNet on the true board. Orders that are illegal on the true board are void. Comparing WM with W2R shows whether success inside the model carries over to the same plan in the engine. The consistency ratio CR $=$ W2R/Real measures the success retained relative to plans made directly in the engine.

Table~\ref{tab:plan-consistency} shows that Task~J has a lower WM success rate than Real for every model. GPT-6-Luna often fails to keep an imagined Task~J trajectory valid to the end, so its WM success is also below W2R. In Task~S, every model has WM above W2R and therefore overestimates the success of the plans made inside it.

\begin{table}[h]
\centering
\small
\setlength{\tabcolsep}{4pt}
\begin{tabular}{@{}llccc@{}}
\toprule
Task & Model & WM & W2R & CR \\
\midrule
J & Qwen3-1.7B (SFT) & 55.8 & 50.8 & 0.78 \\
J & Qwen3-8B (SFT) & 55.6 & 50.6 & 0.78 \\
J & GPT-6-Luna & 18.2 & 34.2 & 0.53 \\
J & GPT-6-Sol & 41.0 & 41.4 & 0.64 \\
\midrule
S & Qwen3-1.7B (SFT) & 82.8 & 41.2 & 0.63 \\
S & Qwen3-8B (SFT) & 82.8 & 42.2 & 0.65 \\
S & GPT-6-Luna & 61.0 & 38.0 & 0.58 \\
S & GPT-6-Sol & 83.6 & 39.8 & 0.61 \\
\bottomrule
\end{tabular}
\caption{Plan consistency on 500 starts. WM is success inside the world model, and W2R is success when its plan is replayed in the engine (both in \%). CR is W2R/Real, where Real $=65.0\%$ is success when the agent acts directly in the engine.}
\label{tab:plan-consistency}
\end{table}

\subsubsection{Consistency with a Different Agent}
\label{app:cr-agent}

We replace the agent's supervised DipNet with an RL-trained DipNet, while the opponents keep the supervised policy. As shown in Table~\ref{tab:cr-agent}, CR changes little in either task. In Task~S, WM is again above W2R.

\begin{table}[h]
\centering
\small
\begin{tabular}{@{}llcccc@{}}
\toprule
Agent & Task & Real & WM & W2R & CR \\
\midrule
Supervised & J & 65.0 & 55.8 & 50.8 & 0.78 \\
 & S & 65.0 & 82.8 & 41.2 & 0.63 \\
\midrule
RL & J & 67.4 & 55.6 & 50.2 & 0.74 \\
 & S & 67.4 & 85.2 & 46.4 & 0.69 \\
\bottomrule
\end{tabular}
\caption{Plan consistency of the Qwen3-1.7B world model (20k) with a supervised or an RL-trained DipNet agent (success rates in \%, 500 starts).}
\label{tab:cr-agent}
\end{table}

\subsection{Choosing among Candidate Actions}
\label{app:candidates}
At each movement phase the agent proposes $B \in \{1,2,4,10\}$ candidate order sets, consisting of DipNet's greedy orders and samples at temperature 0.5. Retreat and adjustment phases are played greedily. A candidate's value counts the centers the agent would own if ownership were updated from the predicted unit positions, plus 0.1 per surviving unit. The agent executes the candidate with the highest value. In Task~J the world model receives the deterministic opponents' actual orders. In Task~S it receives only the agent's candidate. Scoring the same candidates with the engine provides the reference baseline. For Figure~\ref{fig:agent-verifier}b, the Qwen3-1.7B world model trained on 320k transitions is run with $B$ up to 32. We record the chosen candidate's predicted one-phase gain over the current board and its realized gain in the engine.

\begin{table}[h]
\centering
\small
\setlength{\tabcolsep}{3pt}
\begin{tabular}{@{}lrrr@{}}
\toprule
Training data & Real & Imagined & Pairs \\
\midrule
Real 2k & 2,000 & 0 & 12,974 \\
Real 1k + imagined 1k & 1,000 & 1,000 & 12,580 \\
Imagined 2k & 0 & 2,000 & 12,271 \\
\bottomrule
\end{tabular}
\caption{Successful real and imagined games and their board and order pairs for behavior cloning.}
\label{tab:policy-data}
\end{table}

\subsection{Imagined Games and Initialization}
\label{app:policy-learning}
From 2,000 starting boards of the training split, DipNet at temperature 0.1 orders every player for 10 phases in the engine (real games). It also plays the same starts inside the Qwen3-1.7B world model trained on 320k transitions (imagined games). There the world model works in Task~S, so DipNet orders only the agent and the world model imagines the other players. Success is judged by the environment that produced the trajectory. The world model judges 72.5\% of its trajectories successful, against 62.9\% in the engine.

Real 1k and Imagined 1k contain the agent's (board, orders) pairs from 1k successful real or imagined games. The mixture combines 1k of each. To construct Real 2k and Imagined 2k, we use 1,300 additional training starts, disjoint from the first 2,000, and keep enough successful games to reach 2k. The same policy generates the added games in the engine or world model. Table~\ref{tab:policy-data} gives the resulting training-pair counts.

A Qwen3-1.7B policy learns to output the agent's orders from the board and the agent's name. Every trained policy in Table~\ref{tab:agent-bc} uses three epochs, an effective batch size of 32, and a learning rate of $2\times10^{-5}$. Initialized policies train on Real 1k but start from the 20k or 320k Diplomacy world model. Each policy plays on the same 200 test starts. Illegal orders are dropped and their units hold. The behavior-policy row is supervised DipNet, played greedily as the agent.

Paired over 200 test starts, Real 2k gains 0.345 more supply centers than Real 1k (standard error, SE 0.10) and 0.06 more than the mixture (SE 0.11). Imagined 2k gains 0.14 over Imagined 1k (SE 0.12) and is 0.25 below Real 2k (SE 0.11).

\end{document}